\documentclass[journal]{IEEEtran}
\usepackage{times}
\usepackage{epsfig}
\usepackage{graphicx}
\usepackage{amsmath}
\usepackage{amssymb}
\usepackage{lineno}
\usepackage{multirow}
\usepackage{caption}
\usepackage{epstopdf}
\usepackage{subcaption}
\usepackage{hyperref}
\usepackage[table,xcdraw]{xcolor}

\ifCLASSINFOpdf
\else
\fi
\begin{document}
%
\title{Contrast-Oriented Deep Neural Networks for Salient Object Detection}
%
%
%


\author{Guanbin Li and Yizhou Yu

\thanks{This work was supported in part by the National Natural Science Foundation of China under Grant 61702565 and was also sponsored by CCF-Tencent Open Research Fund.}
\thanks{G. Li is with Sun Yat-sen University, Guangzhou 510006, China (e-mail: liguanbin@mail.sysu.edu.cn).}
\thanks{
Y. Yu is with the Department of Computer Science, The University of Hong Kong~(e-mail: yizhouy@acm.org).}

\thanks{A preliminary version of this paper appeared in CVPR 2016~\cite{LiYu16}.}
}

%
%

\markboth{Transactions on Neural Networks and Learning Systems,~Vol.~xx, No.~x, month~2018}%
{Li \MakeLowercase{and} Yu: Contrast-Oriented Deep Neural Networks for Salient Object Detection}
%



\maketitle

\begin{abstract}
Deep convolutional neural networks have become a key element in the recent breakthrough of salient object detection. However, existing CNN-based methods are based on either patch-wise (region-wise) training and inference or fully convolutional networks. Methods in the former category are generally time-consuming due to severe storage and computational redundancies among overlapping patches. To overcome this deficiency, methods in the second category attempt to directly map a raw input image to a predicted dense saliency map in a single network forward pass. Though being very efficient, it is arduous for these methods to detect salient objects of different scales or salient regions with weak semantic information. In this paper, we develop hybrid contrast-oriented deep neural networks to overcome the aforementioned limitations. Each of our deep networks is composed of two complementary components, including a fully convolutional stream for dense prediction and a segment-level spatial pooling stream for sparse saliency inference. We further propose an attentional module that learns weight maps for fusing the two saliency predictions from these two streams. A tailored alternate scheme is designed to train these deep networks by fine-tuning pre-trained baseline models. Finally, a customized fully connected CRF model incorporating a salient contour feature embedding can be optionally applied as a post-processing step to improve spatial coherence and contour positioning in the fused result from these two streams. Extensive experiments on six benchmark datasets demonstrate that our proposed model can significantly outperform the state of the art in terms of all popular evaluation metrics.
\end{abstract}


%
\IEEEpeerreviewmaketitle

\section{Introduction}
Visual saliency detection aims to locate the most conspicuous regions in images according to the human visual system and has recently received increasing research interest. Image saliency detection is traditionally approached in the form of either eye-fixation prediction or salient object detection. The former focuses on the natural mechanism of visual attention and aims at accurately predicting human eye attended image locations. However, previous research has pointed out that salient object detection, which is more concerned with the integrity of the predicted object regions, is more conducive to a series of computer vision tasks including semantic segmentation~\cite{wei2016stc}, object localization and detection~\cite{navalpakkam2006integrated,wang2017multi}, content-aware image editing~\cite{avidan2007seam}, visual tracking~\cite{wu2014weighted} and person re-identification~\cite{bi2014person}. Although numerous valuable models have been proposed, salient object detection remains challenging due to a variety of complex factors in real-world scenarios. 

Perceptual studies~\cite{einhauser2003does,parkhurst2002modeling} have shown that visual contrast is the key factor that affects visual saliency. A series of conventional salient object detection algorithms based on local or global contrast modeling~\cite{cheng2015global,yang2013saliency,wang2013visual} have been successfully proposed. In previous research efforts, visual contrast modeling is generally focused on the differences among various handcrafted low-level features and coupled with heuristic saliency priors. Although handcrafted features tend to perform well in simple cases, they are not robust enough for more challenging scenarios.
For example, it is hard for local contrast models to accurately segment out large homogeneous regions inside salient objects while global contrast information may fail to handle images with cluttered background. Although there exist machine learning based algorithms for salient object detection~\cite{lu2014learning,jiang2013salient,liu2011learning,mai2013saliency}, they are basically focused on integrating various handcrafted features~\cite{jiang2013salient} or merging multiple saliency maps computed by different methods~\cite{mai2013saliency}.

Recently, deep convolutional neural networks have been widely used in salient object detection~\cite{LiYu15,zhao2015saliency,wang2015deep} because of their powerful feature representations and have achieved substantially better performance than traditional methods. Methods based on deep convolutional neural networks can be roughly divided into two categories. Methods in the first category generally perform patch-wise (or region-wise) training and inference. Specifically, an image is first divided into a set of regions or patches, and deep CNN based regression or classification models are then trained to independently map each image patch or region to a saliency score or a binary class label~(salient or non-salient). However, this results in serious storage and computational redundancies, making training and testing very time-consuming. For example, training a patch-oriented CNN model takes over two GPU days while requiring hundreds of megabytes of storage to save deep features extracted from one single image.
Inspired by the latest trends of developing fully convolutional neural networks for pixel-level image understanding problems~\cite{long2014fully,chen2014semantic,xie2015holistically}, methods in the second category train end-to-end models that directly map an input image of arbitrary size to a saliency map with the same size, performing dense feedforward computation and backpropagation over the entire image. This type of methods have rapidly become the cornerstone of this field as they not only achieve very favorable performance but also are very efficient. However, it is still arduous for these methods to detect salient objects of different scales or salient regions with weak semantic information. Moreover, pixel-level correlation is typically not considered in such fully convolutional networks (FCNs), which usually give rise to incomplete salient regions with blurry contours.



In this work, we develop hybrid contrast-oriented deep neural networks to overcome the aforementioned limitations of two types of contemporary CNN-based salient object detection methods. 
Our deep networks are composed of a fully convolutional stream for dense prediction and a segment-level spatial pooling stream for sparse saliency inference. We devise a multi-scale fully convolutional network (MS-FCN) in the first stream, which receives an entire image as input and directly learns to map it to a dense saliency prediction with pixel-level accuracy. Our MS-FCN can not only learn multi-scale feature representations, but also accurately judge the saliency of every pixel by mining visual contrast information hidden in multi-scale receptive fields. 
The segment-level spatial pooling stream computes another sparse saliency map over superpixels by modeling the contrast between every superpixel and its spatially adjacent regions. It extracts multi-scale regional features very efficiently by performing feature masking in the feature map of an intermediate layer of MS-FCN. At the end, we produce our final saliency map by merging the saliency maps from both streams with weight maps generated from a proposed attentional module in our deep network. Our MS-FCN can also be re-trained to generate a contour map for salient objects. This contour map can be used to improve contour localization in the fused saliency map via a fully connected CRF.

In summary, this paper has the following contributions:
\begin{itemize}
\item We propose end-to-end contrast-oriented deep neural networks for localizing salient objects using multi-scale contextual information. They incorporate a fully convolutional stream for dense prediction and a segment-wise spatial pooling stream for sparse inference.
    A tailored alternate scheme is designed to train these deep networks by fine-tuning pre-trained baseline models. 
\item A multi-scale VGG-16 or ResNet-101 network pre-trained for image classification is re-purposed as the fully convolutional stream to infer a dense saliency prediction directly from the raw input image in a single forward pass. This fully convolutional network can also be re-trained to infer a salient object contour map, which can be represented as a feature embedding and incorporated in a fully connected CRF model to further improve contour localization in the final result.
\item We have also devised a segment-wise spatial pooling stream complementary to the fully convolutional stream in our deep network. This stream efficiently masks out segment-wise features from one designated feature map of MS-FCN, and accurately models visual contrast among superpixels and well captures saliency discontinuities along region boundaries.
\end{itemize}

The rest of this paper is organized as follows. Section~\ref{sec:relatedwork} reviews related work on salient object detection. In Section~\ref{sec:deep_contrast_network}, we introduce our proposed contrast-oriented deep neural networks. The complete algorithm is presented in Section~\ref{sec:algo}. Section~\ref{sec:experiment} provides extensive performance evaluation as well as comparisons against state-of-the-art models. Finally, we conclude this paper in Section~\ref{sec:conclusion}.

\section{Related Work}\label{sec:relatedwork}

Traditional salient object detection can be categorized into bottom-up approaches with handcrafted low-level features~~\cite{gao2007bottom,achanta2009frequency,liu2011learning,klein2011center,perazzi2012saliency,yang2013saliency,jiang2013salient,zhu2014saliency,cheng2015global,wang2013saliency} and top-down approaches incorporating high-level knowledge~\cite{judd2009learning,chang2011fusing,goferman2012context,shen2012unified,liu2014adaptive,jia2013category,li2014secrets}. Bottom-up methods are usually based on the center bias or background priors and infer saliency maps from global or local contrast represented as a combination of handcrafted low-level features~(e.g. color, texture and image gradient). 
Bottom-up computational models are primarily based on a center-surround scheme and compute saliency maps using a linear or non-linear combination of low-level features such as color, intensity, texture and orientation of edges~\cite{achanta2009frequency,cheng2015global,hou2007saliency,liu2011learning}. Top-down methods are in general task-dependent and require a machine learning scheme to incorporate high-level knowledge into a process which was originally limited to specified objects or assumptions~\cite{jia2013category,li2014secrets,liu2014adaptive}. Graph based methods have also been widely used to enhance spatial consistency and refine detected saliency maps~\cite{lei2016universal,yang2013saliency,LiYu16}.
Recently, deep learning based methods have been widely used for salient object detection and have promoted its research into a new phase. 
Since the focus of this paper is deep learning based salient object detection, we highlight the most relevant previous work in the following discussion.

In recent years, the successful application of deep convolutional neural networks has triggered a revolution in machine learning and artificial intelligence, and has yielded significant improvement in a variety of visual comprehension tasks, including image classification~\cite{krizhevsky2012imagenet}, object detection~\cite{girshick2014rich} and semantic segmentation~\cite{long2014fully}, closing the gap to human-level performance.
Motivated by this, several attempts have also been made to apply deep neural network models to salient object detection~\cite{li2016visual,LiYu16,li2015deepsaliency,li2017instance,LiAAAI18}. Han {\em et al.}~\cite{han2016two} first attempted to develop stacked denoising autoencoders to learn powerful representations for salient object detection in an unsupervised and bottom-up manner. In~\cite{li2015weighted}, a weighted sparse coding framework is proposed for image saliency detection. Recently, with the widespread application of convolutional neural networks in image analysis and comprehension tasks, it is not surprising to see a surging number of research papers where very good results have been achieved on salient object detection via the application of CNNs. 
Li {\em et al.}~\cite{LiYu15,li2016visual} trained a multi-layer fully connected network for deriving the saliency value of every superpixel from its contextual CNN features. Wang {\em et al.}~\cite{wang2015deep} proposed two deep neural networks, which take into account both low-level features and high-level objectness, for salient object detection at the patch level. 
A multi-context deep CNN framework incorporating both global and local contexts is presented in~\cite{zhao2015saliency}. 
However, all these methods include fully connected layers and infer saliency maps in an isolated patch-wise manner, the crucial spatial information in the input image is ignored. However, since all the image patches are treated as independent samples during network training and inference, there is no shared computation among overlapping image segments, which results in significant redundancies and excessive computational cost during training and testing.

To address these issues, inspired by the seminal work of developing end-to-end deep networks for semantic image segmentation~\cite{long2014fully,chen2014semantic}, variant of fully convolutional neural networks have been introduced to solve the problem of salient object detection since the publication of our earlier conference version~\cite{LiYu16}.
Li {\em et al.}~\cite{li2015deepsaliency} proposed to explore the correlations between saliency detection and semantic image segmentation using a multi-task fully convolutional neural network. Liu {\em et al.}~\cite{liu2016dhsnet} propose a hierarchical recurrent CNN to progressively refine the details of saliency maps from a coarse prediction result generated from the forward pass of a fully convolutional VGG-16 network. Kuen {\em et al.}~\cite{kuen2016recurrent} proposed a recurrent attentional convolution-deconvolution network (RACDNN), which consists of a recurrent neural network and a spatial transform module, to recurrently attend to selected image sub-regions for saliency refinement. 
In~\cite{wang2016saliency}, Wang {\em et al.} introduced a recurrent fully convolutional network (RFCN) to iteratively refine the saliency map with incorporated prior knowledge. 
These FCN based models have greatly improved both accuracy and efficiency in saliency detection, there are still three aspects of the flaws. First of all, these models are mostly based on the topmost feature map of the network for saliency inference, the over-reliance on the regional semantic feature may result in the pool detection performance on the salient region with weak semantic information. Second, all of these methods consider feature modeling at a single scale and may not accurately detect salient objects of very different sizes. And finally, as the value at each position of a saliency map generated from FCN-based models is derived from a context with a fixed size~(receptive field), the contours of salient objects can hardly be well detected, and the generated saliency maps usually have inadequate spatial consistency. 
Our proposed method instead delves into the nature of saliency prediction, capturing the key aspect in this problem, which is contrast learning. The proposed method is not only able to infer a saliency probability map from the contrast information in a multiscale deep CNN but also from edge-preserving region-wise contrast information. In addition, it has been proven that fully connected CRFs can be formulated as recurrent neural networks (RNNs). However, experimental results show that RNNs can hardly be trained to achieve comparable results as CRFs. Our proposed method therefore exploits the effectiveness of a contour-aware CRF.
Our experimental results demonstrate the superiority of our proposed method in comparison to all existing FCN based salient object detection techniques.

Note that the initial deep contrast network reported in CVPR 2016~\cite{LiYu16} can be viewed as the first piece of work that aims at designing an end-to-end fully convolutional network for visual contrast modeling. To a certain extent, it inspired the subsequent development of FCN-based models in this field. Our updated contrast-oriented deep neural network for salient object detection has several improvements over its initial version. First, we adapt the state-of-the-art ResNet-101 network~\cite{he2015deep} for image classification to a fully convolutional network and use it to replace the VGG-16 network in the original fully convolutional stream, achieving better performance. Second, the fully convolutional stream is run on multiple scaled versions of the original input image while the segment-wise spatial pooling stream is trained using segments from multi-level image segmentation. These strategies make our deep model more accurately detect salient objects at different scales. Third, we propose to add an attentional module which learns pixel-wise soft weights for fusing the two saliency maps respectively generated from the two streams. Fourth, we discover that the proposed multi-scale fully convolutional stream in our deep network can be re-trained to detect salient region contours, which can be integrated into a fully connected CRF model to further improve contour localization in the final saliency map. Finally, we present a more comprehensive experimental comparison among multiple model variants and report improved results on all benchmarks using all evaluation metrics.

\begin{figure*}[ht]
\begin{center}
   \includegraphics[width=0.95\textwidth]{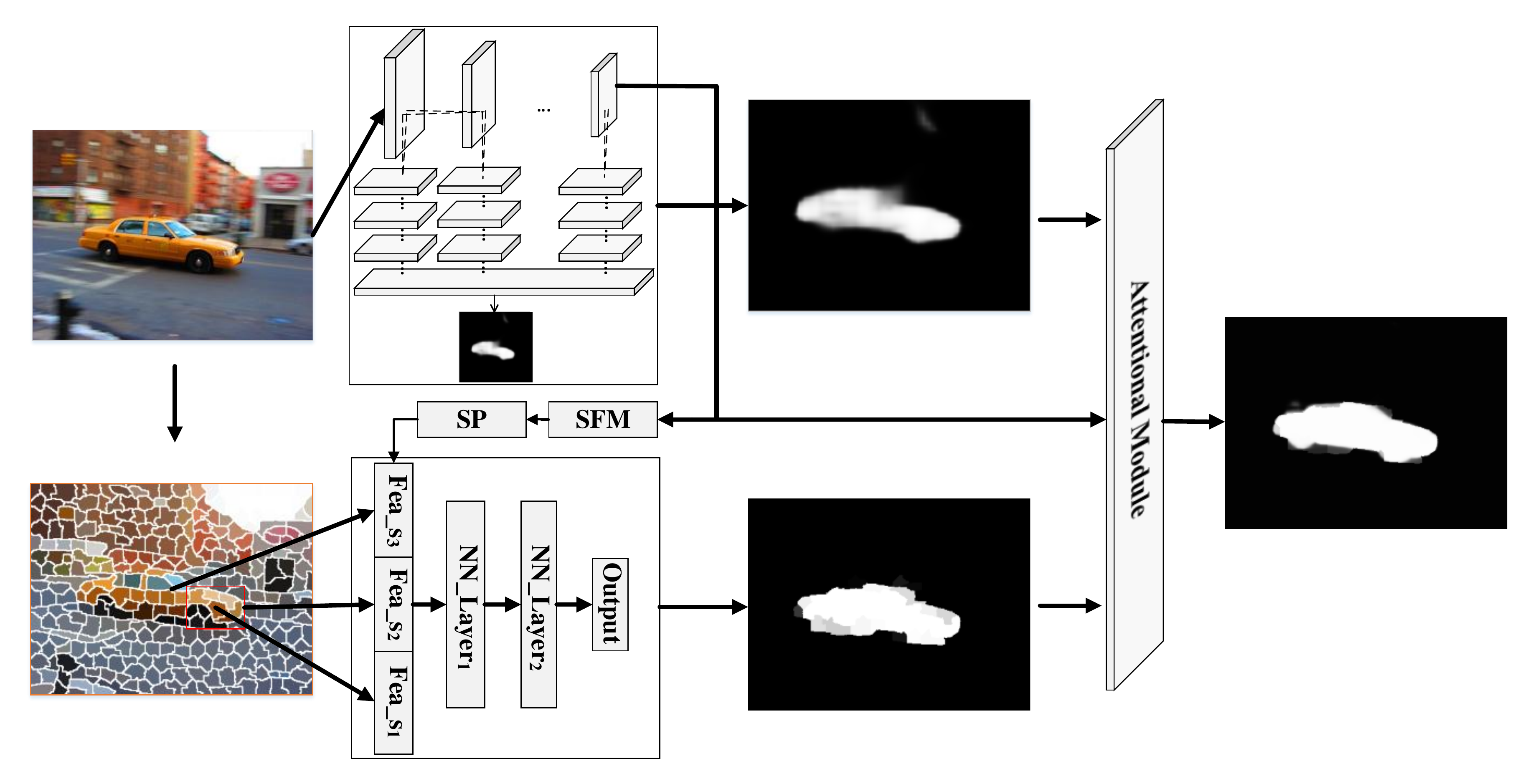}
\end{center}
   \caption{The overall architecture of our proposed contrast-oriented deep neural network. It consists of a fully convolutional stream~(upper part), a segment-wise spatial pooling stream~(lower part) and an attentional module to fuse the intermediate saliency maps from the two streams. ``SFM'' refers to the segment feature masking layer while ``SP'' refers to the spatial pooling operation.}
\label{fig:dcn}
\end{figure*}

\section{Deep Contrast Network}\label{sec:deep_contrast_network}
As illustrated in Fig.~\ref{fig:dcn}, our proposed contrast-oriented deep neural network is composed of two complementary components, a fully convolutional stream for dense saliency prediction and a segment-wise spatial pooling stream for sparse saliency inference. Specifically, the first component is a multi-scale fully convolutional network~(MS-FCN), which receives an entire image as input and is trained to map the input to a dense saliency map $S_1$ in an end-to-end mode by exploiting visual contrast across multiple levels of feature maps. The segment-wise spatial pooling stream is trained to infer the saliency map $S_2$ at the segment level by discovering the contrast among spatially adjacent regions on the basis of features masked out from one designated feature map of the first stream and a multi-layer perceptron. At the end, these two intermediate saliency predictions from the above two network streams are merged according to weight maps prescribed by a trained attention module. The merged map becomes our final saliency map $S$.

\subsection{Multi-Scale Fully Convolutional Network}\label{sec:ms-fcn}
Inspired by the groundbreaking application of fully convolutional networks in pixel-level image comprehension, we focus on constructing an end-to-end pixelwise regression network, which can directly map a raw input image to a dense saliency map. Considering the centrality of contrast modeling for saliency detection, we have the following considerations when designing the structure of this end-to-end network. First, the network should be deep enough to accommodate features from multiple levels since visual saliency relies on modeling the contrast among both low-level appearance features as well as high-level semantic features. Second, the network needs to be able to explore the visual contrast across multiple feature maps and detect salient objects of various scales. Finally, due to the lack of training images with pixel-wise labeling, it is much desired to fine-tune an existing pre-trained network instead of training from scratch.


As VGG~\cite{simonyan2014very} and ResNet~\cite{he2015deep} are the two most representative and widely used deep classification networks with publicly available pre-trained models, we choose them as our pre-trained networks and adapt for our requirements. Here we describe in detail the transformation of the VGG-16 network, and ResNet-101 can be similarly transformed to satisfy the requirements. To re-purpose the VGG-16 network for dense saliency map generation, we first convert the two fully connected layers of VGG-16 into $1\times 1$ convolutional ones as described in~\cite{long2014fully}. Moreover, as the original VGG-16 network consists of 5 max pooling layers and each with stride 2, the resulting network can only yield low-resolution prediction maps with $1/32$ the input resolution. To make the resulting saliency map have a higher resolution, we remove the downsampling operation in the last two max-pooling layers by simply setting their ``stride'' to 1, which results in downsampling by a factor of 8 instead of 32. At the same time, to maintain the same size of the receptive fields of the convolutional layers that follow, we refer to~\cite{chen2014semantic,li2014highly} and apply the dilation operation to the corresponding filter kernels. The dilation algorithm~(also called \`a trous algorithm), which was originally proposed to improve the computational efficiency of undecimated wavelet transforms~\cite{mallat1999wavelet}, has recently been incorporated into the Caffe framework~\cite{chen2014semantic,li2014highly} as ``dilated convolution'' to efficiently control the resolution of feature maps within deep CNNs without the need to learn extra parameters. It works by inserting zeros between filter weights. Specifically, 
consider applying the dilated version of a convolutional filter $w$ to an input feature map $x$, and generating an output feature map $y$. The output value at position $i$ is calculated as
\begin{equation}
  y[i]=\sum_{k}{x[i+r\cdot k]w[k]},
\end{equation}
where the dilation rate $r$ corresponds to the stride with which we sample the input feature map. This is equivalent to applying convolution to the input feature map $x$ with filters up-sampled by inserting $r-1$ zeros between any two originally adjacent filter elements along each dimension. This dilated convolution allows us to explicitly control the density of feature responses in our customized fully convolutional networks. In our implementation, after setting the stride of the last two pooling layers to $1$, we replace all subsequent convolutional layers with dilated convolutional layers with dilation rate $r=2$ or $r=4$~($r=2$ for the three consecutive convolutional layers after the penultimate max-pooling layer and $r=4$ for the last two newly converted $1\times 1$ convolutional layers).

\begin{figure*}[ht]
\begin{center}
   \includegraphics[width=0.475\textwidth
   ]{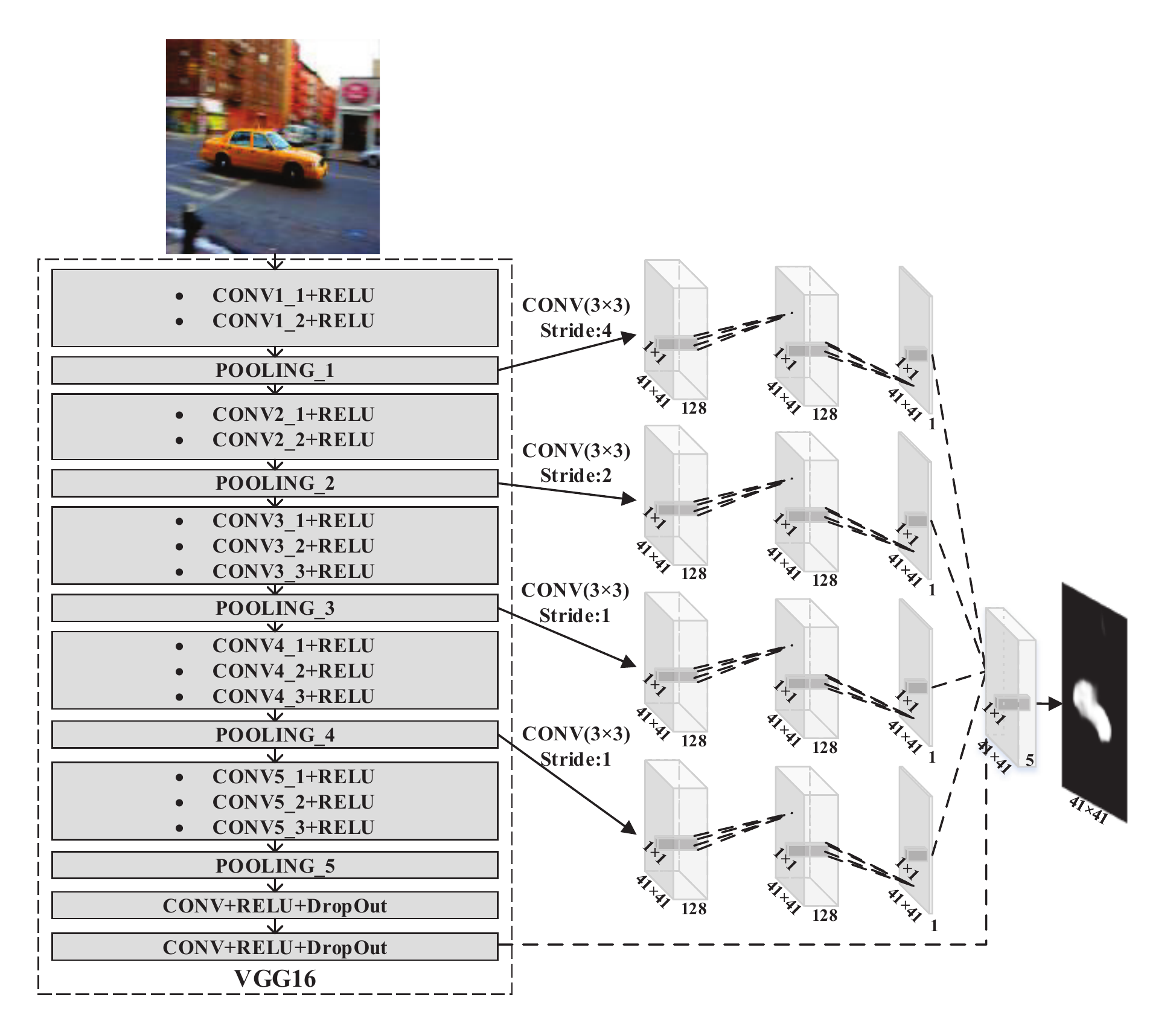}
    \includegraphics[width=0.475\textwidth
   ]{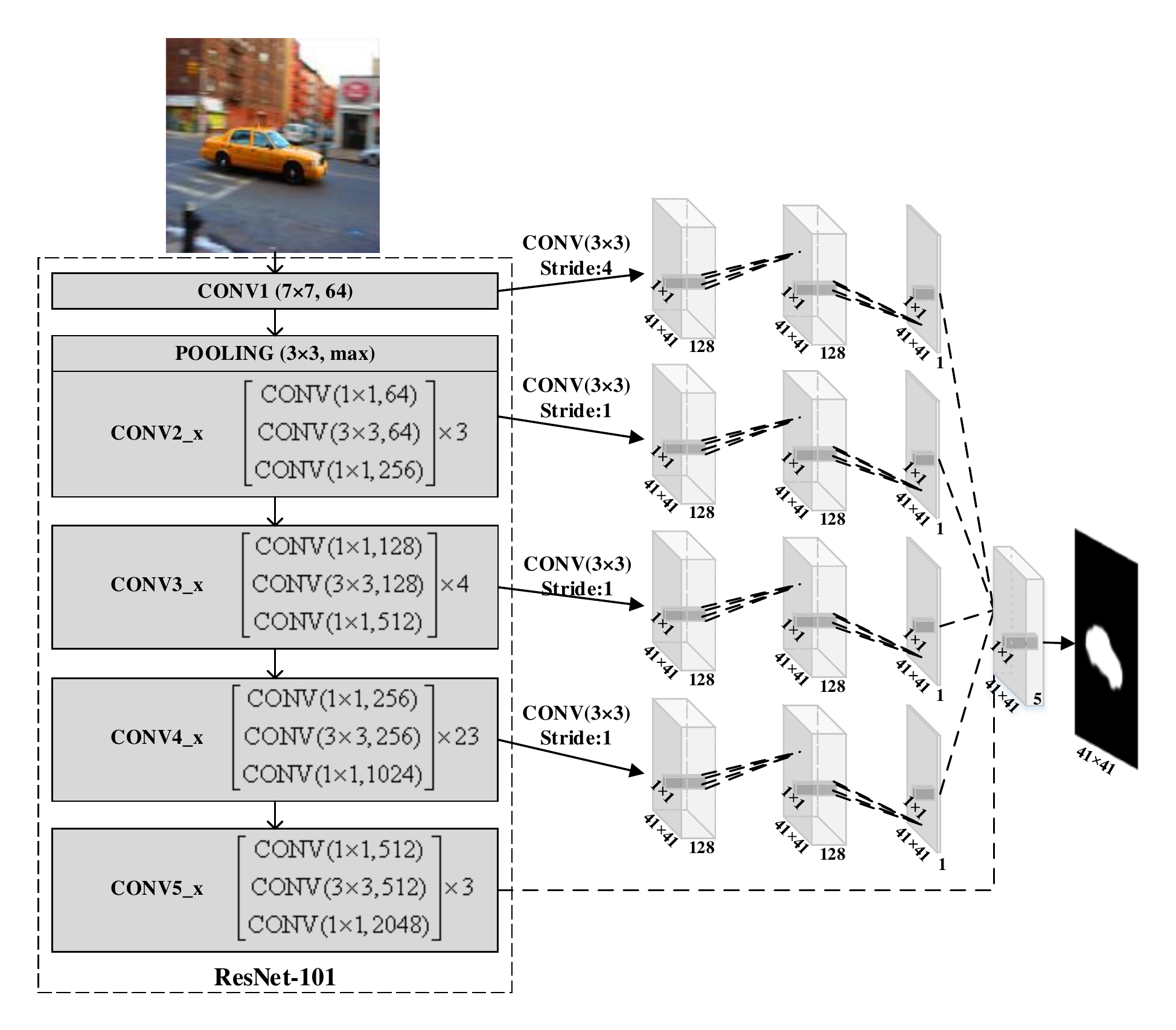}
\end{center}
   \caption{The architecture of VGG-16 based multi-scale fully convolutional network~(left) and ResNet-101 based multi-scale fully convolutional network~(right). We connect three extra convolutional layers to each of the first four max-pooling layers of VGG-16 and convert it to a multiscale version. For ResNet-101, we divide the 101 layers into five groups and connect an extra sub-network with three convolutional layers to each of the final layers in the first four groups to form the multiscale version.}
\label{fig:ms-fcn}
\end{figure*}

VGG-16 has five max pooling layers performing downsampling operations. If we start from the pooling layer closest to the input image, these pooling layers have an increasingly larger receptive field containing contextual information. To design a deep convolutional network that is capable of mining visual contrast information crucial in saliency inference, we further develop a multiscale network from the above fully convolutional version of VGG-16. As shown in the left part of Fig.~\ref{fig:ms-fcn}, we connect three extra convolution layers to each of the first four max-pooling layers. The first extra layer uses $3\times 3$ convolution kernels and has 128 channels while the second one uses $1\times 1$ convolution kernels and also has 128 channels. And the third extra layer has one $1\times 1$ kernel and a single channel, which is used to produce the output saliency map. To make the output feature maps of the four sets of extra convolutional layers have the same size (8$\times$ downsampling resolution), the stride of the first layer in these four sets are set to 4, 2, 1, and 1, respectively. Although the four resulted feature maps are of the same size, they are computed using receptive fields with different sizes and hence represent contextual features at 4 different scales. We further stack these four feature maps with the last output feature map of the above customized fully convolutional conversion. The stacked feature maps (5 channels) are fed into a final convolution layer with a $1\times 1$ kernel and a single output channel, which is modulated by the sigmoid activation function to produce the saliency probability map. Though the resulting saliency map of this network stream has a downsampling factor of $8$ in comparison to the input image, it is smooth enough and allows us to use simple bilinear interpolation to restore the resolution of the original input at a negligible computational cost.
We call this resized saliency map $S_1$.

Note that the ResNet-101 network has no hidden fully connected layers. To adapt ResNet-101 for dense saliency prediction, we simply replace its 1000-way linear classification layer with a linear convolutional layer with a $1\times 1$ kernel and a single output channel. Similar to VGG-16, the resolution of the feature maps before the linear convolutional layer is only $1/32$ of that of the original input image because the original ResNet-101 consists of one pooling layer and $4$ convolutional layers, each of which has stride $2$. We call these five layers ``down-sampling layers''. As described in~\cite{he2015deep}, the $101$ layers in ResNet-101 can be divided into five groups. Feature maps computed by different layers in each group share the same resolution. To increase the resolution of the final saliency map, we replace the last two down-sampling layers with dilated convolution layers, skip subsampling by setting their stride to $1$, and correspondingly increase the dilation rate of subsequent convolution kernels to enlarge their receptive fields. Therefore, all the features maps in the last three groups have the same resolution, $1/8$ original resolution, after network transformation. To develop a multiscale version of the above end-to-end extension of ResNet-101, as shown in the right of Fig.~\ref{fig:ms-fcn}, we connect an extra sub-network with three convolutional layers to each of the final layers in the first four groups. These additional layers have the same structure as those added to VGG-16. Similar to the multiscale extension of VGG-16, the four output feature maps from these four sub-networks are stacked together with the final output feature map of the transformed ResNet-101, and fed into a final convolutional layer with a $1\times 1$ kernel and a single output channel for final saliency map inference.

\subsection{Segment-Level Saliency Inference}
Salient objects in images are usually presented in a variety of irregular shapes and the corresponding saliency map often exhibits discontinuities along the object boundaries. Our multiscale fully convolutional network operates at a subsampled pixel level and equally treats each pixel in the input image without explicitly taking into account such saliency discontinuities. To better model visual contrast between regions and visual saliency along region boundaries, we design a segment-wise spatial pooling stream in our network.

We first divide an input image into a set of superpixels, and call each superpixel a segment. A mask is computed for every segment in the feature map generated from one selected convolutional layer of MS-FCN, which is named the feature masking layer.
We choose the convolutional layer Conv5\_3 as the feature masking layer in the MS-FCN based on VGG-16, and the last convolutional layer in the fourth layer group as the feature masking layer in the MS-FCN based on ResNet-101 as suggested in~\cite{he2015deep}. Since the activations at each location in the feature masking layer is controlled by a receptive field in the input image, we first project every location in the feature masking layer to the center of its receptive field as in~\cite{girshickICCV15fastrcnn}. For each segment in the input image, we first generate a binary mask within its bounding box. In this mask, pixels inside the segment are labeled `$1$' while others are labeled `$0$'. Each pixel labeled as `$1$' in the binary mask is first assigned to the closest receptive field center and then backprojected onto the feature masking layer. Thus, each location in the feature masking layer collects multiple `$1$' labels backprojected from its receptive field. The ratio between the number of collected `$1$' labels at the location and the number of pixels in the input image closest to its receptive field center is recorded. To yield a binary mask for the segment on the feature masking layer, the previously computed ratio at every location is thresholded at 0.5, and the set of locations with nonzero values after thresholding form the segment mask. In the event that the ratio at all locations is below 0.5, the set of locations with nonzero ratios before thresholding form the segment mask. The resulting segment mask is then applied to the output feature map of the feature masking layer by simply multiplying this binary mask with each channel of the feature map. We call the resulting features segment-masking features in our method.
Note that the feature map generated from the feature masking layer has a downsampling factor of 8 instead of 32 in the original VGG-16 network or 16 in the original ResNet-101 network since subsampling has been skipped in the last two downsampling layers as described in Section~\ref{sec:ms-fcn}. Therefore, the resolution of the feature map generated from the feature masking layer is sufficient for segment masking.

Since segments have irregular shapes and variable sizes when projected onto the feature masking layer, we further perform a spatial pooling (SP) operation to produce a feature vector of fixed length for each segment. It is a simplified version of spatial pyramid pooling described in~\cite{he2014spatial}. Specifically, we divide the bounding box of a projected segment into~$h\times w$~cells and perform Max- or mean-pooling over valid positions~(with mask label `$1$') in each grid cell. This results in $h\times w$ feature vectors of size $C$, which is the number of convolutional filters in the feature masking layer. Afterwards, we concatenate the feature vectors extracted from all grid cells of the same segment to obtain the final feature vector with $h\times w\times C$ dimensions for that segment.

To discover segment-level visual contrast, we represent each segment with a concatenation of three feature vectors respectively for three nested and increasingly larger regions masked out from the designated feature map. These three regions include the bounding box of the considered segment, the bounding box of the immediate neighboring segments as well as the entire feature map from the feature masking layer~(with the considered segment excluded to indicate the position of the segment). The above-mentioned feature representation of each segment is further fed into two fully connected layers. The output of the second fully connected layer is fed into a ``Sigmoid'' layer which employs the sigmoid function to perform logistic regression and produces a distribution over binary saliency labels. We call the saliency map generated in this way $S_2$.

In fact, this segment-wise spatial pooling stream of our network is an accelerated version of our previous work proposed in~\cite{LiYu15}. Although they share the identical idea of inferring saliency from contrast among multiscale contextual regions, feature extraction and processing in the current method is much more efficient as hundreds of segmental features for the same image are instantaneously masked out from the feature map generated by MS-FCN in a single forward pass. 
Moreover, our segment-wise spatial pooling stream also achieves better results as segment features are extracted from our multiscale fully convolutional network, which has been fine-tuned for salient object detection, instead of from the original VGG-16 model for image classification.

\subsection{Attentional Module for Saliency Map Fusion}\label{sec:attentional_module}
To merge predicted saliency scores from the two different streams, there are three straightforward options: average pooling, max-pooling and $1 \times 1$ convolution. However, all these strategies are image content independent. As our two network streams have complementary strengths in saliency map prediction, inspired by~\cite{bahdanau2014neural,chen2015attention}, we design a trainable attentional module to generate content-dependent weight maps for fusing the results from the two streams.

Let $S_1$ and $S_2$ be the probabilistic saliency maps from the two network streams, respectively. The final saliency map from our deep contrast network is calculated as a weighted sum of these two maps. The spatially varying weights are adaptively learned. Therefore, they are called weight maps. 
Let $S$ be the fused saliency map, $W_1$ be the weight map for the saliency map generated from the MS-FCN stream and $W_2$ be the weight map for the saliency map generated from the second stream. The merged saliency map is calculated by summing the element-wise product between each probability map~(resized to $1/8$ the input image resolution) and its corresponding weight map:
\begin{equation}
  S = W_1\odot S_1 + W_2 \odot S_2.
\end{equation}

We refer to~\cite{chen2015attention} and call $W_1$ and $W_2$ attention weights as they reflect how much attention should be paid to individual network streams as well as saliency scores at different spatial locations. These two attention weights can also be considered as feature maps that have the same size as the predicted saliency maps, and thus can be jointly trained in a fully convolutional network. In this work, we employ a differentiable attention module to our deep network to infer these attention weights. 
As illustrated in Fig.~\ref{fig:dcn}, the proposed attention module receives as input the output feature map from the feature masking layer, and it contains two convolutional layers. The first layer has 512 filters with kernel size $3\times 3$ while the second layer has two convolutional filters with kernel size $1\times 1$. The output feature map has two channels, further fed into a SoftMax layer, which generates two score maps corresponding to the aforementioned two attention weights.

\subsection{Deep Contrast Network Training}\label{sec:dcn_train}
We propose an alternate training scheme to train our network. Specifically, in the initialization phase, we pre-compute the segments of all training images and train the segment-wise spatial pooling stream alone until convergence to obtain its initial network parameters. Segment-wise saliency labeling is performed by thresholding the average pixel-wise labeling inside each segment, and the segment features are extracted using the VGG-16 or ResNet-101 image classification model pre-trained on the ImageNet dataset~\cite{deng2009imagenet}. After initialization, we alternately update the weights in the two network streams. First, we fix the weights of the second stream and train the MS-FCN as well as the attention module for one epoch. Note that the weights in the attention module for adaptively merging the predicted saliency maps from the two streams are trained simultaneously with the MS-FCN stream in an end-to-end mode. Next, we fix the weights in the MS-FCN as well as the attention module, and fine-tune the parameters in the second stream for one epoch using segment features extracted with an updated VGG-16 or ResNet-101 network embedded in the MS-FCN stream. We alternately train the two streams 8 times~(16 epochs in total) until the whole training process converges.
We define the following class-balanced cross-entropy as the loss function for training the multi-scale fully convolutional steam and the attention module of our network,
\begin{equation}
\begin{aligned}
L=&-\beta_i\sum_{i=1}^{|I|}G_i\log P\left ( S_i=1|I_i,W \right ) \\& -  \left ( 1-\beta_i \right )\sum_{i=1}^{|I|}\left ( 1-G_i \right )\log P\left ( S_i=0|I_i,W \right ),
\end{aligned}
\end{equation}
where $\beta_i$ represents the class balancing weight, denoted as $\beta_i=\frac{|I|\_}{|I|}$ and $1-\beta_i=\frac{|I|_+}{|I|}$, where $|I|$, $|I|_+$ and $|I|\_$ respectively indicate the total number of pixels, salient pixels and non-salient ones in image $I$. $G$ represents the groundtruth annotation and $W$ represents the collection of all network weights in the MS-FCN stream and the attention module.
When fine-tuning the segment-wise spatial pooling stream, we use a batch of images as a unit and update parameters by minimizing the summed squared errors accumulated over all segments from the same batch of training images.


\section{The Complete Algorithm}\label{sec:algo}

\subsection{Superpixel Segmentation}\label{sec:superpixel}
The segment-wise spatial pooling stream of our network requires the input image to be decomposed into non-overlapping segments. In order to better avoid artificial boundaries in the generated saliency map, each segment should be a perceptually homogeneous region while at the same time, strong contours and edges should still be well preserved. In our earlier version~\cite{LiYu16}, we use a geodesic distance~\cite{criminisi2010geodesic} based SLIC algorithm for superpixel generation. In this work, 
we discover that graph based image segmentation~\cite{felzenszwalb2004efficient} produces segments with better edge preservation than the SLIC algorithm, and using segments generated from multiple levels of image segmentation can further improve the performance. Therefore, we refer to~\cite{felzenszwalb2004efficient} and employ the graph based image segmentation algorithm therein to generate three levels of segments with different parameter settings. 
We train a single segment-wise spatial pooling stream for all the segments across three levels of segmentation instead of learning different model parameters for segments from different levels of segmentation. When generating a saliency map from the segment-wise spatial pooling stream, we apply the same stream to infer a saliency map for each level of segmentation and then simply average the three resulting saliency maps.


\subsection{Salient Contour Detection}\label{sec:saliency_contour}
While in most cases our proposed deep contrast network works well, it sometimes produces saliency maps where salient region boundaries are not accurately localized, particularly for images containing small salient regions. Meanwhile, we find that our multi-scale fully convolutional network described in Section~\ref{sec:ms-fcn}, when re-trained using annotated salient region contours, is also capable of detecting the contours of salient regions. The detected contours can be further encoded as feature vectors and embedded into a CRF framework to enhance spatial coherence and the preservation of salient region contours in saliency maps. 
To prepare training data for salient region contour detection, boundary pixels of salient regions in the groundtruth saliency maps are labeled `$1$', and all other pixels are labeled `$0$'. Such salient region contour maps are taken as the groundtruth annotations when MS-FCN is trained for salient region contour detection, and the class-balancing weight is updated according to the fraction of pixels on salient region contours.

Given a detected salient region contour map $M$, we apply the normalized cut~\cite{shi2000normalized} algorithm to generate per-pixel feature vectors, which are used in a fully connected CRF to improve boundary localization in our final saliency map. First, we construct a sparse graph where every pixel is connected to other pixels in its $11\times 11$ neighborhood. The affinity matrix $W$ of this graph is defined as follows,
\begin{equation}
W_{ij}=\exp \left( -\max_{p\in \overline{ij}} \left\{ \frac{M(p)^2}{\rho} \right\} \right),
\end{equation}
where $W_{ij}$ denotes the affinity between pixels $i$ and $j$, $p$ represents pixels along the line segment ($\overline{ij}$) connecting pixels $i$ and $j$, $M(p)$ indicates the probability of pixel $p$ being on a salient region contour, and $\rho$ is a constant scaling factor, which is set to 0.1 in our experiments. The idea is that two pixels should have a similar saliency value if there is no salient region contour crossing the line segment connecting these two pixels. Given an affinity matrix $W$, we further define $D_{ii} = \Sigma_{i\neq j}W_{ij}$, and solve for generalized eigenvectors of the following system, $\left(D-W\right)v= \lambda Dv$.
We use these eigenvectors as additional features to improve spatial coherence. In our experiments, we use eigenvectors corresponding to the 16 smallest eigenvalues.


\begin{figure}[b]
\begin{center}
   \includegraphics[width=0.95\columnwidth]{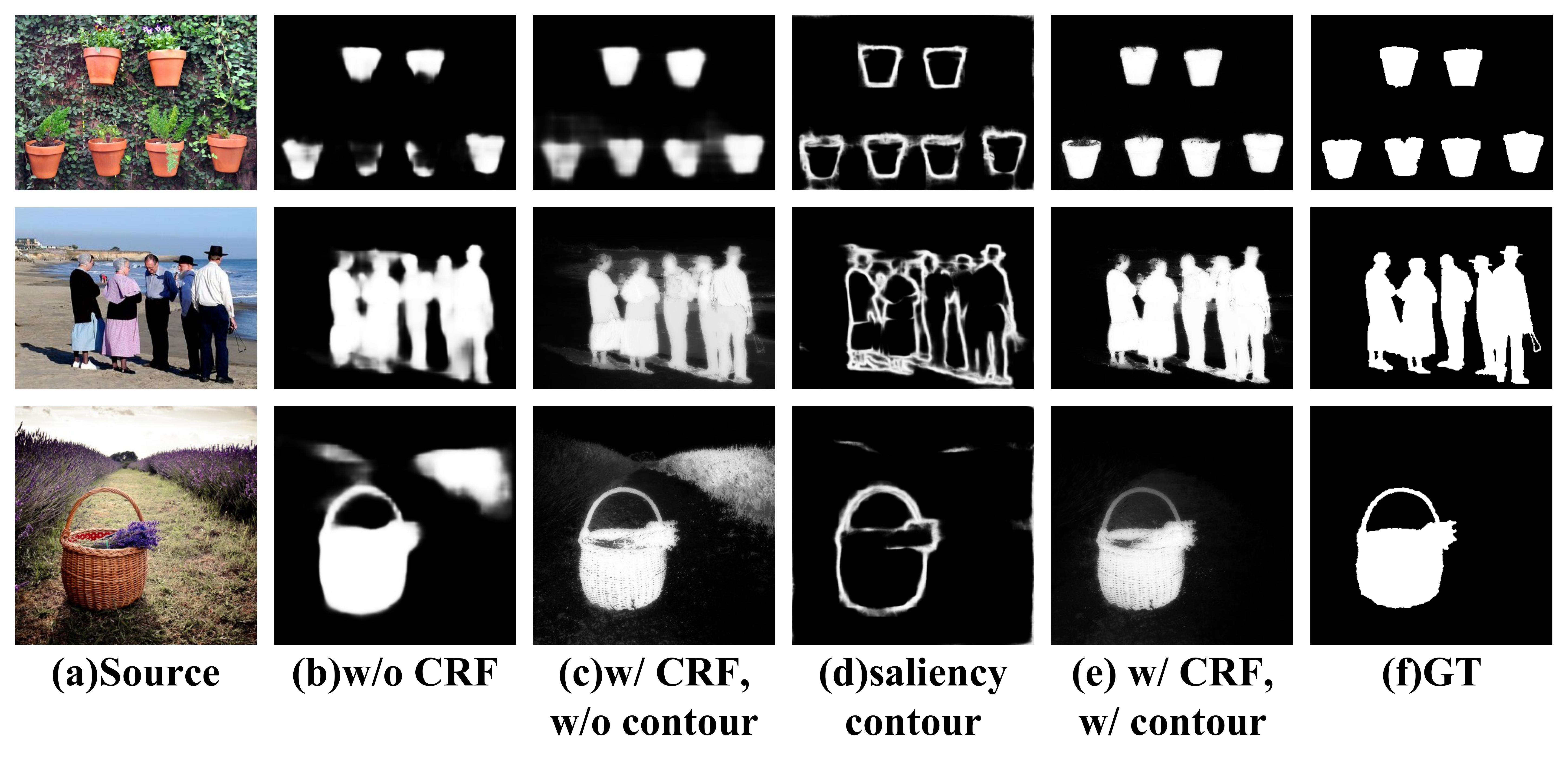}
\end{center}
   \caption{Examples of saliency maps generated with and without a CRF~(including CRFs with and without a contour feature embedding).}
\label{fig:crf_effect}
\end{figure}

\subsection{Spatial Coherence}\label{sec:spatial_coherence}
Since both streams of our deep contrast network independently infer the saliency score of each individual pixel or segment without considering the impact of the correlation among pixels and segments on saliency prediction,
the resulting saliency maps contain more or less incomplete or false positive salient objects. To mitigate this issue, we adopt a fully connected conditional random field (CRF)~\cite{krahenbuhl2012efficient} in a post-processing step to enhance spatial coherence. The energy function of the CRF model is formulated as
\begin{equation}
E\left( L \right) = -\sum_{i}\log P\left(l_i\right)+\sum_{i,j}\theta_{ij}\left(l_i, l_j\right),
\end{equation}
where $L$ is the binary label prediction for all pixels~(salient or not salient). $P(l_i)$ indicates the probability of pixel $x_i$ being labeled $l_i$.
As an initialization, $P(l_i=1)=S_i$ and $P(l_i=0)=1-S_i$, where $S_i$ refers to the predicted probabilistic saliency value at pixel $x_i$ of the saliency map $S$ generated from our deep contrast network. The pairwise potential $\theta_{ij}\left(l_i,l_j\right)$ is defined as
\begin{equation}
\begin{split}
\theta_{ij}=\mu\left(l_i,l_j\right)\Bigg[ \omega_1\exp\Bigg(-\frac{\left \|p_i-p_j  \right \|^2}{2\sigma_\alpha^2}-\frac{\left \|I_i-I_j \right \|^2}{2\sigma_\beta^2}\\ - \frac{\left \|v_i-v_j \right \|^2}{2\sigma_\gamma ^2}\Bigg) + \omega_2\exp\left(-\frac{\left \|p_i-p_j \right\|^2}{2\sigma_\epsilon^2}\right)\Bigg],
\end{split}
\end{equation}
where $\mu\left(l_i,l_j\right) = 1$ if $l_i \neq l_j$, and zero, otherwise. It involves a summation of two Gaussian kernels. The first kernel is based on the observation that neighboring pixels should be assigned similar saliency scores if they have similar colors but do not have intervening salient region contours. It therefore depends on pixel positions~($p$), pixel intensities~($I$) and the contour feature embedding ($v$) discussed in Section~\ref{sec:saliency_contour}. The importance of color similarity, spatial closeness and salient region contours are controlled by three parameters ($\sigma_\alpha$, $\sigma_\beta$ and $\sigma_\gamma$), respectively. 
The second kernel is only dependent on pixel positions with hyperparameter $\sigma_\epsilon$ controlling the scale of the Gaussian function. As pointed out in~\cite{shotton2009textonboost}, it helps to enhance label smoothness and remove small isolated regions.

As it has been proved in~\cite{krahenbuhl2012efficient}, this energy minimization process can be modeled as efficient approximate probabilistic inference by adopting a mean-field approximation to the original CRF. High-dimensional filtering can be employed to speed up the computation.
We adapt the publicly available implementation of~\cite{krahenbuhl2012efficient} to minimize the above energy function. The optimization process takes less than 0.5 second for an image with $300\times 400$ pixels. After CRF model optimization, a saliency map $S_{crf}$ can be generated from the pixelwise posterior probabilities of saliency labels. We visualize the effectiveness of our CRF in Fig.~\ref{fig:crf_effect}. As can be seen, the original saliency maps from the proposed method without CRF are rather coarse and the integrity~(spatial coherence) of detected salient regions can hardly be maintained. 
Though saliency maps generated with a traditional CRF~(without the contour feature embedding) can enhance the spatial coherence of detected salient regions to some extent, salient region contours still may not be well positioned and there may be false detections in the smooth background~(e.g.~the third row). The fourth column of the figure demonstrates salient region contours detected by our proposed method. As can be seen, it is usually possible to accurately capture the boundaries of salient regions and its corresponding embedded features can further enhance the consistency of saliency prediction across salient region contours and correct prediction errors. A quantitative analysis of our CRF based saliency refinement will be provided in Section~\ref{sec:effectiveness_spatial_coherence}.



\begin{figure*}[ht]
\begin{center}
   \includegraphics[width=0.95\textwidth]{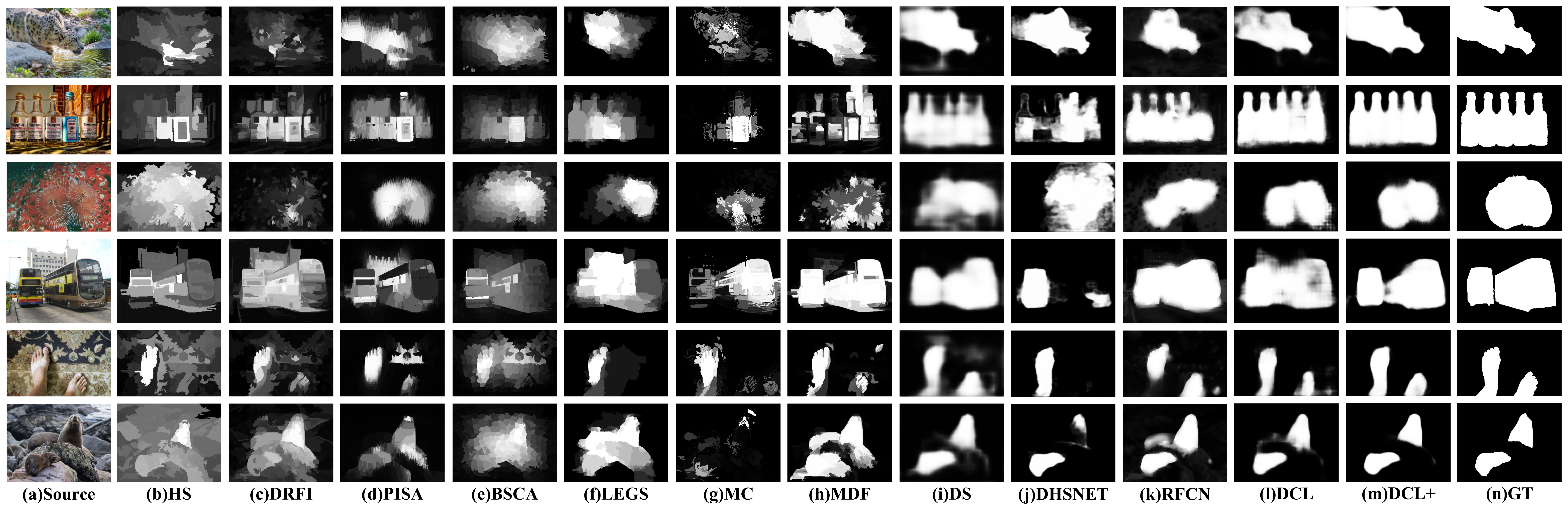}
\end{center}
   \caption{Visual comparison between our methods~(DCL and DCL$^+$) and other state-of-the-art methods. Source: the input images; GT: ground truth saliency maps; DCL$^+$: DCL with CRF refinement. DCL$^+$ consistently achieves the best results in a variety of complex scenarios.
   }
\label{fig:smaps}
\end{figure*}

\begin{table*}[]
\centering
\resizebox{0.95\textwidth}{!}
{
\begin{tabular}{|c|c|c|c|c|c|c|c|c|c|c|c|c|c|c|c|}
\hline
Data Set                    & Metric & SF    & GC    & HS    & DRFI  & PISA  & BSCA                         & LEGS                         & MC                                    & MDF                          & DS                                    & RFCN                                  & DHSNet                                & DCL                                   & DCL+                                  \\ \hline
                            & maxF   & 0.700 & 0.719 & 0.813 & 0.845 & 0.837 & 0.830                        & {\color[HTML]{333333} 0.870} & {\color[HTML]{333333} 0.894}          & 0.885                        & {\color[HTML]{32CB00} \textbf{0.898}} & ---                                   & ---                                   & {\color[HTML]{3166FF} \textbf{0.929}} & {\color[HTML]{FE0000} \textbf{0.931}} \\ \cline{2-16}
\multirow{-2}{*}{MSRA-B}    & MAE    & 0.166 & 0.159 & 0.161 & 0.112 & 0.102 & 0.130                        & {\color[HTML]{333333} 0.081} & {\color[HTML]{32CB00} \textbf{0.054}} & 0.066                        & 0.067                                 & ---                                   & ---                                   & {\color[HTML]{3166FF} \textbf{0.046}} & {\color[HTML]{FE0000} \textbf{0.042}} \\ \hline
                            & maxF   & 0.548 & 0.597 & 0.727 & 0.782 & 0.764 & 0.758                        & {\color[HTML]{333333} 0.827} & 0.837                                 & 0.832                        & {\color[HTML]{333333} 0.900}          & 0.899                                 & {\color[HTML]{32CB00} \textbf{0.907}} & {\color[HTML]{3166FF} \textbf{0.921}} & {\color[HTML]{FE0000} \textbf{0.925}} \\ \cline{2-16}
\multirow{-2}{*}{ECSSD}     & MAE    & 0.219 & 0.233 & 0.228 & 0.170 & 0.150 & 0.183                        & {\color[HTML]{333333} 0.118} & {\color[HTML]{333333} 0.100}          & 0.105                        & 0.079                                 & 0.091                                 & {\color[HTML]{3166FF} \textbf{0.059}} & {\color[HTML]{32CB00} \textbf{0.061}} & {\color[HTML]{FE0000} \textbf{0.058}} \\ \hline
                            & maxF   & 0.590 & 0.588 & 0.710 & 0.776 & 0.753 & 0.723                        & {\color[HTML]{333333} 0.770} & 0.798                                 & {\color[HTML]{333333} 0.861} & {\color[HTML]{333333} 0.866}          & {\color[HTML]{32CB00} \textbf{0.896}} & 0.892                                 & {\color[HTML]{3166FF} \textbf{0.909}} & {\color[HTML]{FE0000} \textbf{0.913}} \\ \cline{2-16}
\multirow{-2}{*}{HKU-IS}    & MAE    & 0.173 & 0.211 & 0.213 & 0.167 & 0.127 & 0.174                        & {\color[HTML]{333333} 0.118} & 0.102                                 & {\color[HTML]{333333} 0.076} & 0.079                                 & 0.073                                 & {\color[HTML]{32CB00} \textbf{0.052}} & {\color[HTML]{3166FF} \textbf{0.050}} & {\color[HTML]{FE0000} \textbf{0.041}} \\ \hline
                            & maxF   & 0.495 & 0.495 & 0.616 & 0.664 & 0.630 & 0.617                        & {\color[HTML]{333333} 0.669} & {\color[HTML]{333333} 0.703}          & {\color[HTML]{333333} 0.694} & {\color[HTML]{32CB00} \textbf{0.773}} & 0.747                                 & ---                                   & {\color[HTML]{3166FF} \textbf{0.799}} & {\color[HTML]{FE0000} \textbf{0.811}} \\ \cline{2-16}
\multirow{-2}{*}{DUT-OMRON} & MAE    & 0.147 & 0.218 & 0.227 & 0.150 & 0.141 & 0.191                        & {\color[HTML]{333333} 0.133} & {\color[HTML]{333333} 0.088}          & {\color[HTML]{333333} 0.092} & {\color[HTML]{32CB00} \textbf{0.084}} & 0.095                                 & ---                                   & {\color[HTML]{3166FF} \textbf{0.070}} & {\color[HTML]{FE0000} \textbf{0.064}} \\ \hline
                            & maxF   & 0.493 & 0.539 & 0.641 & 0.690 & 0.660 & {\color[HTML]{333333} 0.666} & {\color[HTML]{333333} 0.752} & 0.740                                 & 0.764                        & {\color[HTML]{32CB00} \textbf{0.834}} & 0.832                                 & 0.824                                 & {\color[HTML]{3166FF} \textbf{0.851}} & {\color[HTML]{FE0000} \textbf{0.857}} \\ \cline{2-16}
\multirow{-2}{*}{PASCAL-S}  & MAE    & 0.240 & 0.266 & 0.264 & 0.210 & 0.196 & 0.224                        & {\color[HTML]{333333} 0.157} & 0.145                                 & 0.145                        & {\color[HTML]{333333} 0.108}          & 0.118                                 & {\color[HTML]{3166FF} \textbf{0.094}} & {\color[HTML]{32CB00} \textbf{0.098}} & {\color[HTML]{FE0000} \textbf{0.092}} \\ \hline
                            & maxF   & 0.516 & 0.526 & 0.646 & 0.699 & 0.660 & {\color[HTML]{333333} 0.654} & {\color[HTML]{333333} 0.732} & 0.727                                 & 0.785                        & {\color[HTML]{32CB00} \textbf{0.829}} & 0.805                                 & 0.823                                 & {\color[HTML]{3166FF} \textbf{0.848}} & {\color[HTML]{FE0000} \textbf{0.857}} \\ \cline{2-16}
\multirow{-2}{*}{SOD}       & MAE    & 0.267 & 0.284 & 0.283 & 0.223 & 0.223 & 0.251                        & {\color[HTML]{333333} 0.195} & 0.179                                 & {\color[HTML]{333333} 0.155} & {\color[HTML]{32CB00} \textbf{0.127}} & 0.161                                 & {\color[HTML]{32CB00} \textbf{0.127}} & {\color[HTML]{3166FF} \textbf{0.122}} & {\color[HTML]{FE0000} \textbf{0.120}} \\ \hline
\end{tabular}
}
\caption{Quantitative comparison in terms of maximum F-measure (larger is better) and MAE (smaller is better). The three best performing algorithms are marked in \color[HTML]{FE0000}\textbf{red}\color{black}, \color[HTML]{3166FF}\textbf{blue}\color{black}, and \color[HTML]{32CB00}\textbf{green}\color{black}, respectively. As the testing set of the MSRA-B dataset is used as part of the training set in the released model of DHSNet~\cite{liu2016dhsnet} and RFCN~\cite{wang2016saliency}, and the part of the DUT-OMRON dataset is also used in training the DHSNet model, we exclude the corresponding results here.
}
\label{tab:comp_quantity}

\end{table*}

\begin{figure*}[t]
    \centerline{
    \includegraphics[width = 0.237\textwidth]{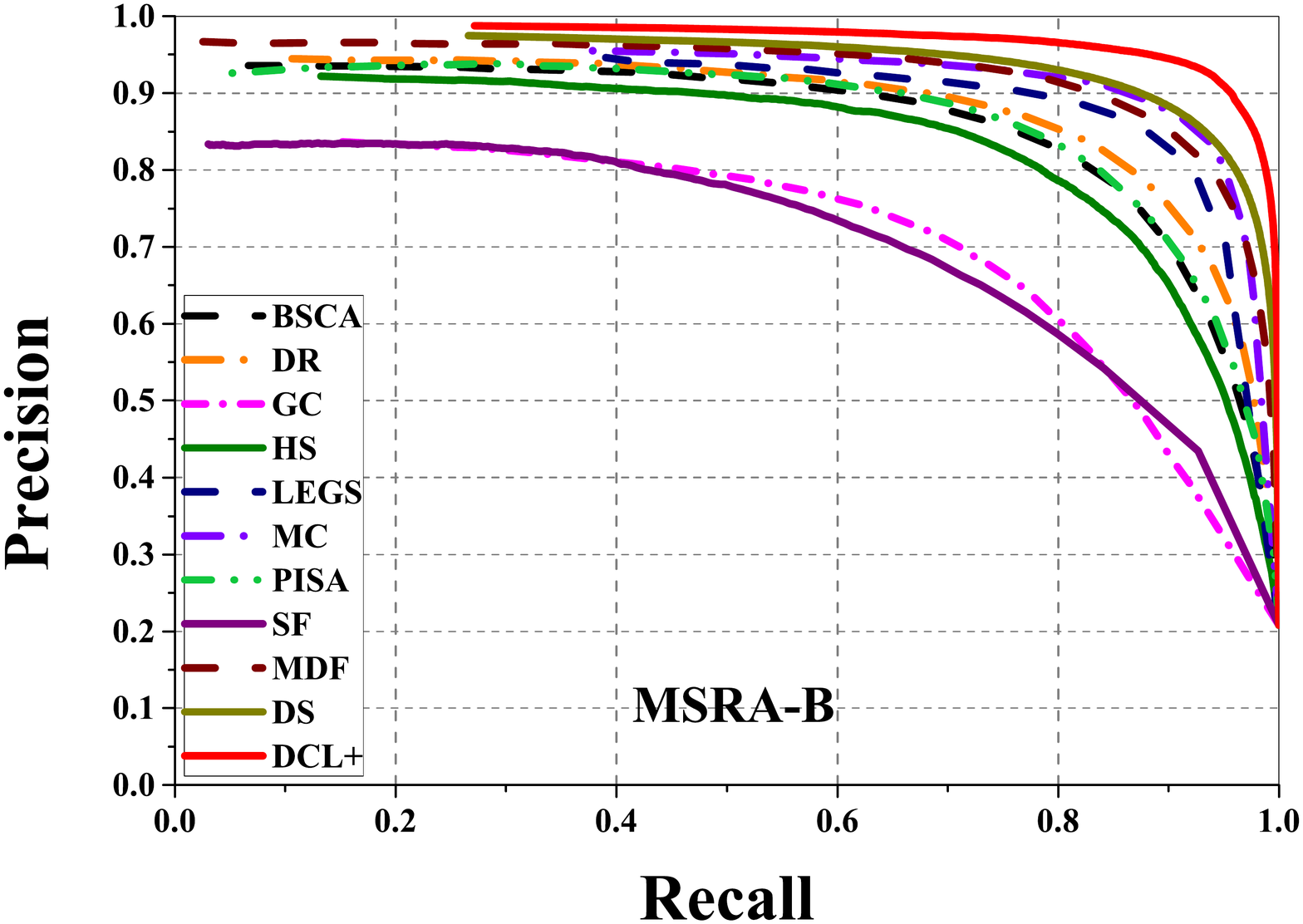}\hfill
    \includegraphics[width = 0.237\textwidth]{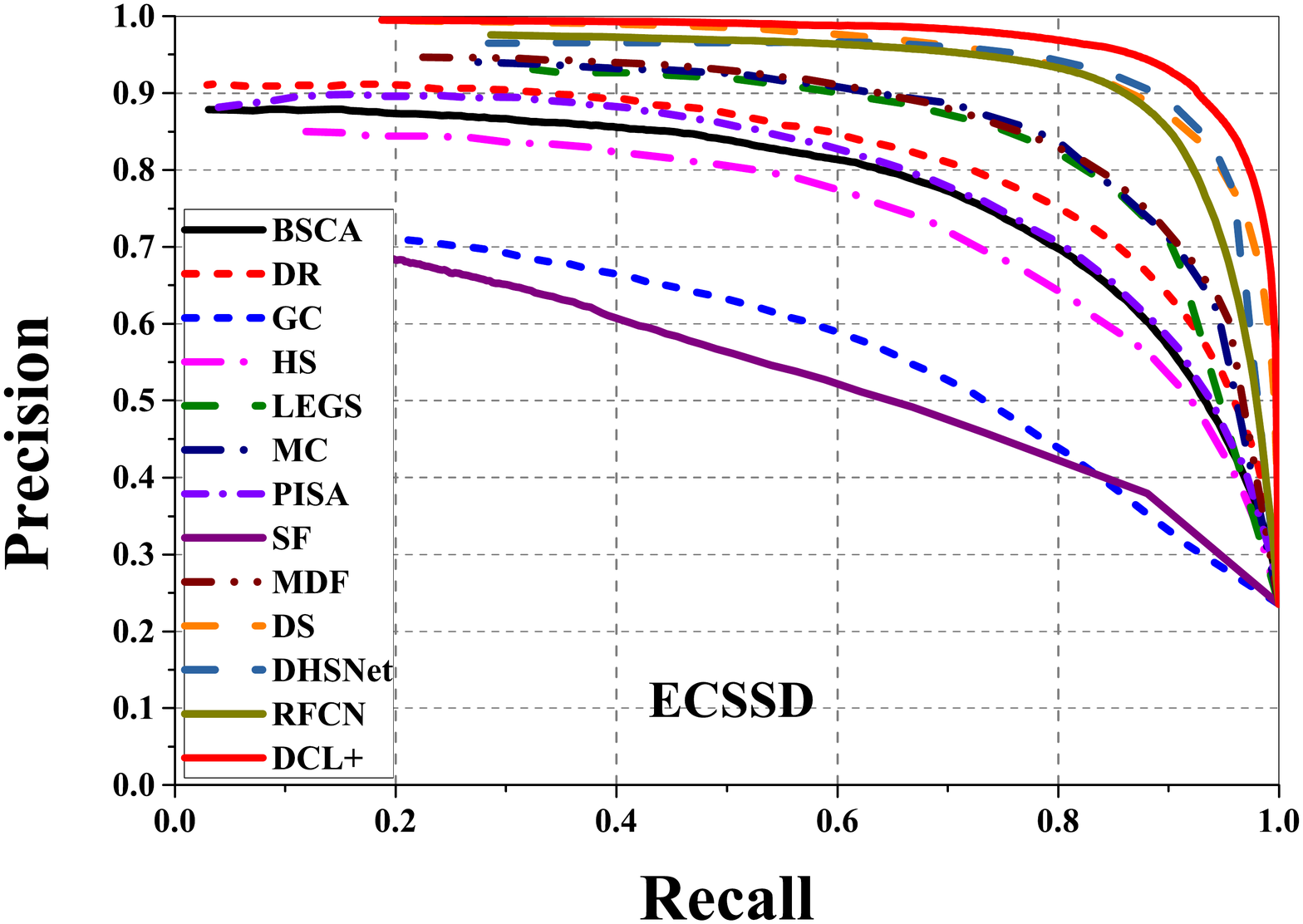}\hfill
    \includegraphics[width = 0.237\textwidth]{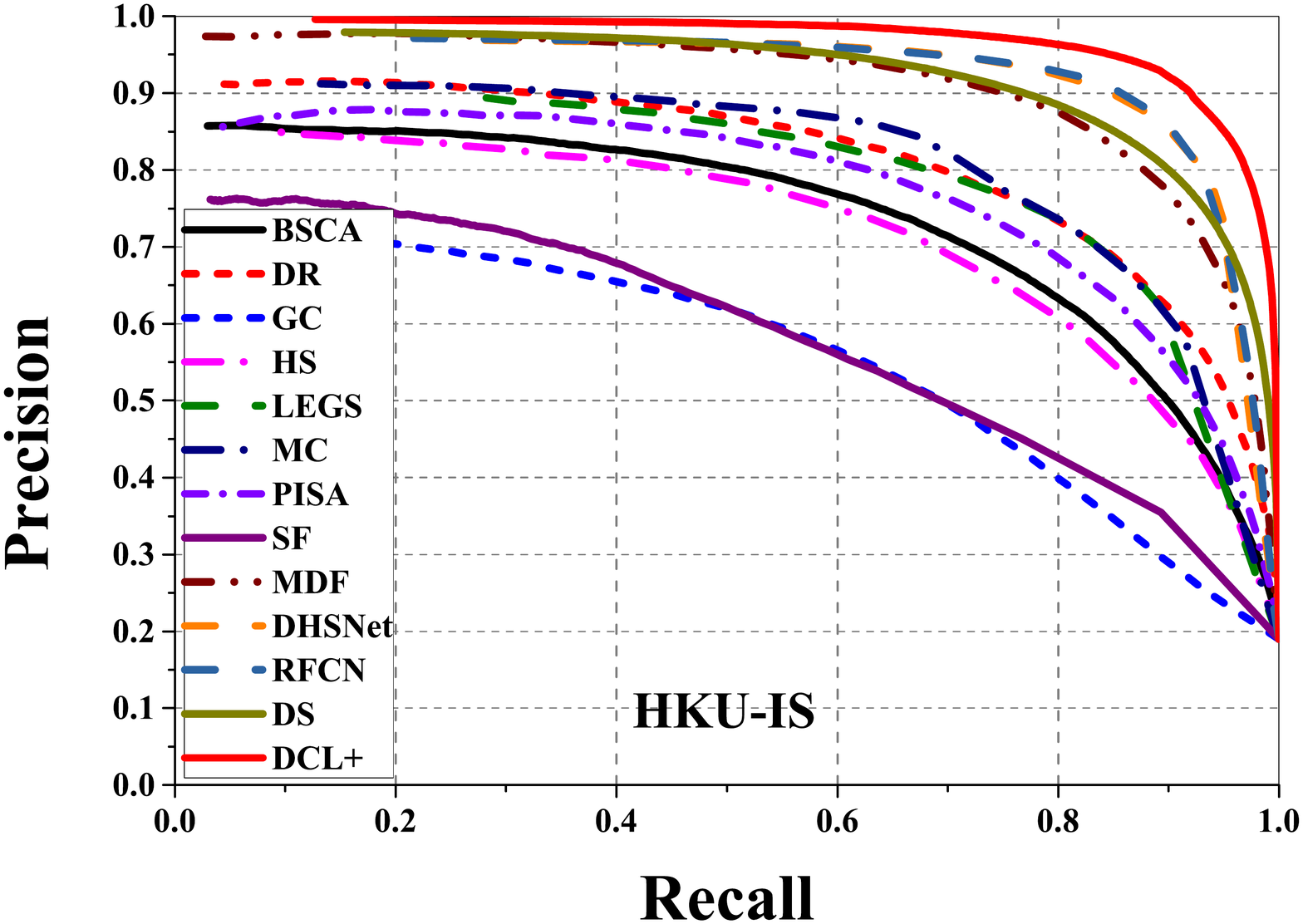}\hfill
    \includegraphics[width = 0.237\textwidth]{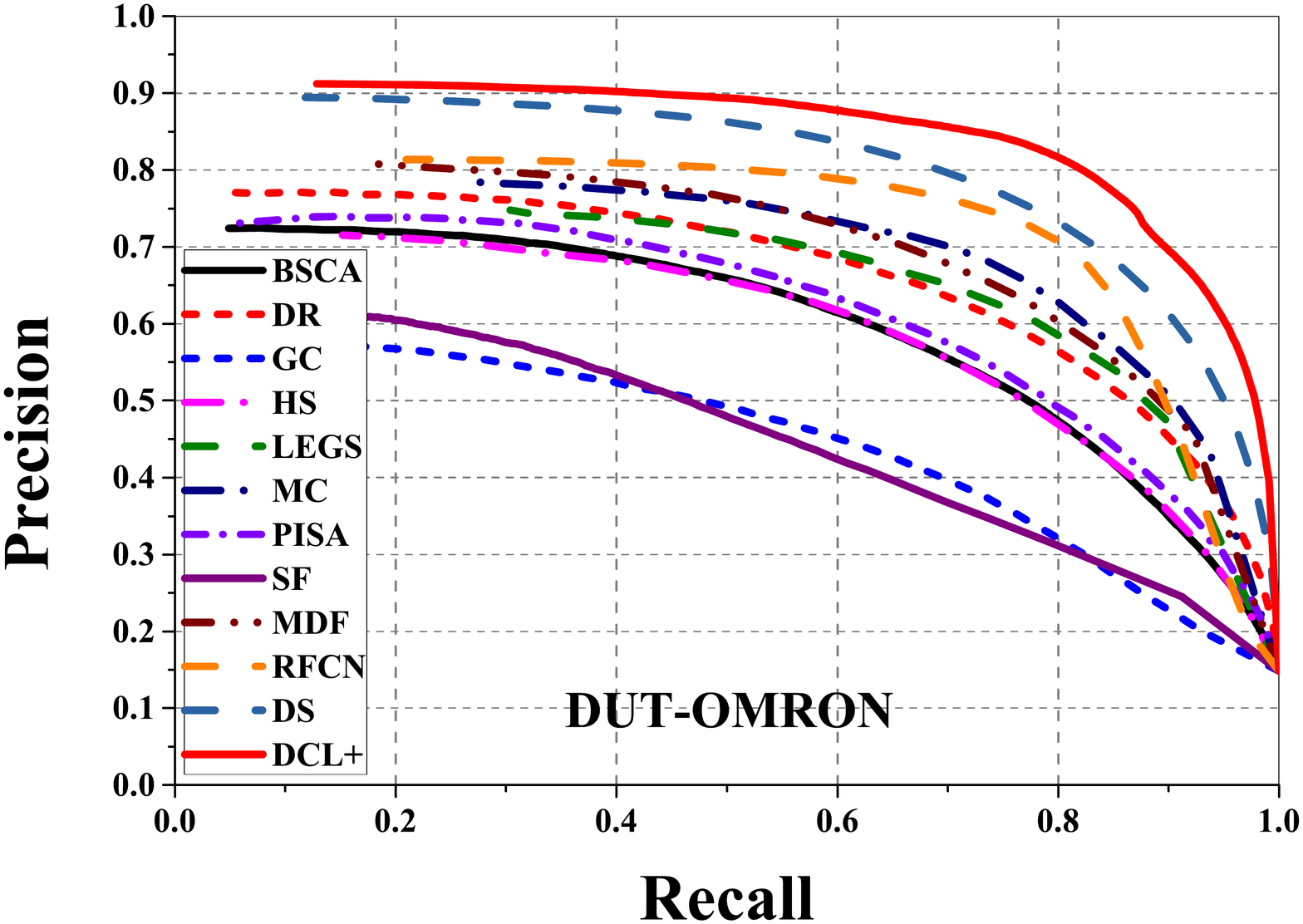}
  }
  \caption{Precision-recall curves of our method and 12 other state-of-the-art algorithms on 4 benchmark datasets. Our DCL$^+$~(DCL with CRF) consistently performs better than other methods across all the benchmarks. 
  }
  \label{fig:comps_pr}
\end{figure*}

\begin{figure*}[t]
    \centerline{
    \includegraphics[width = 0.237\textwidth]{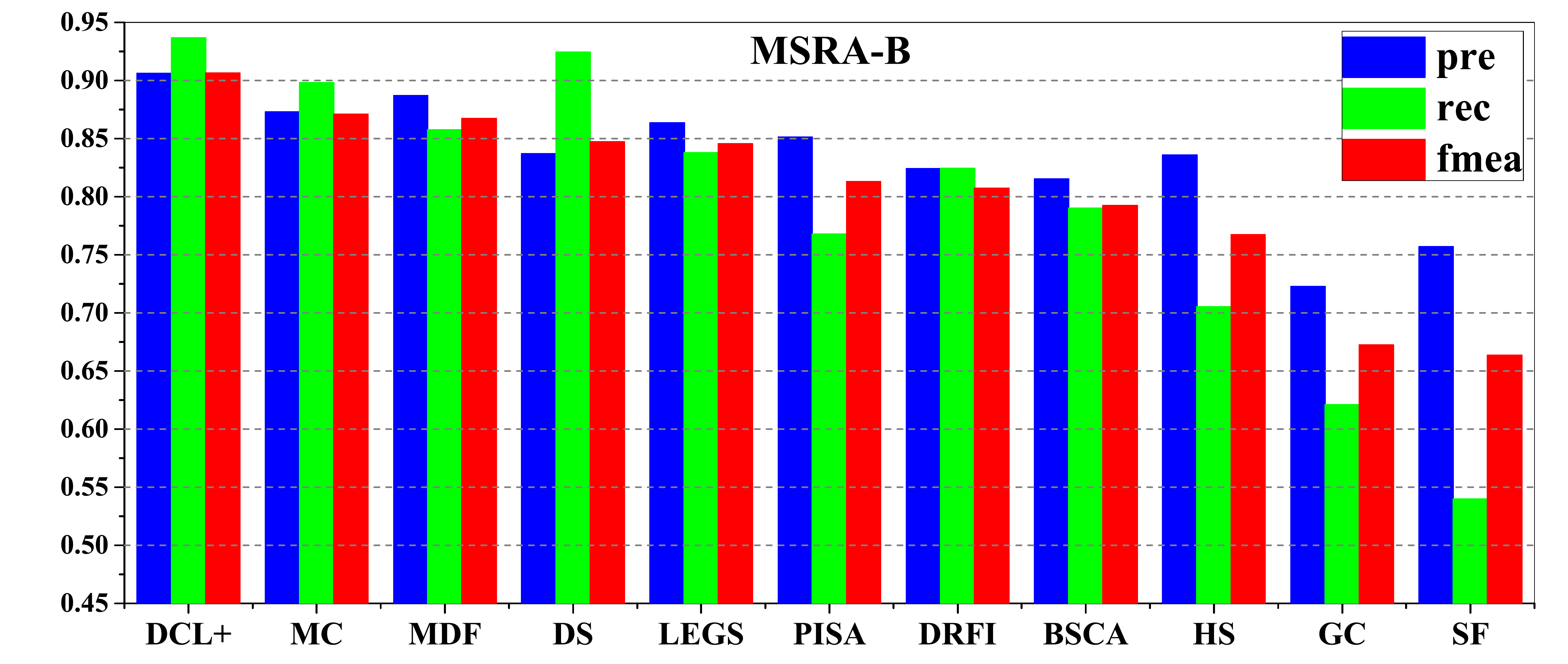}\hfill
    \includegraphics[width = 0.237\textwidth]{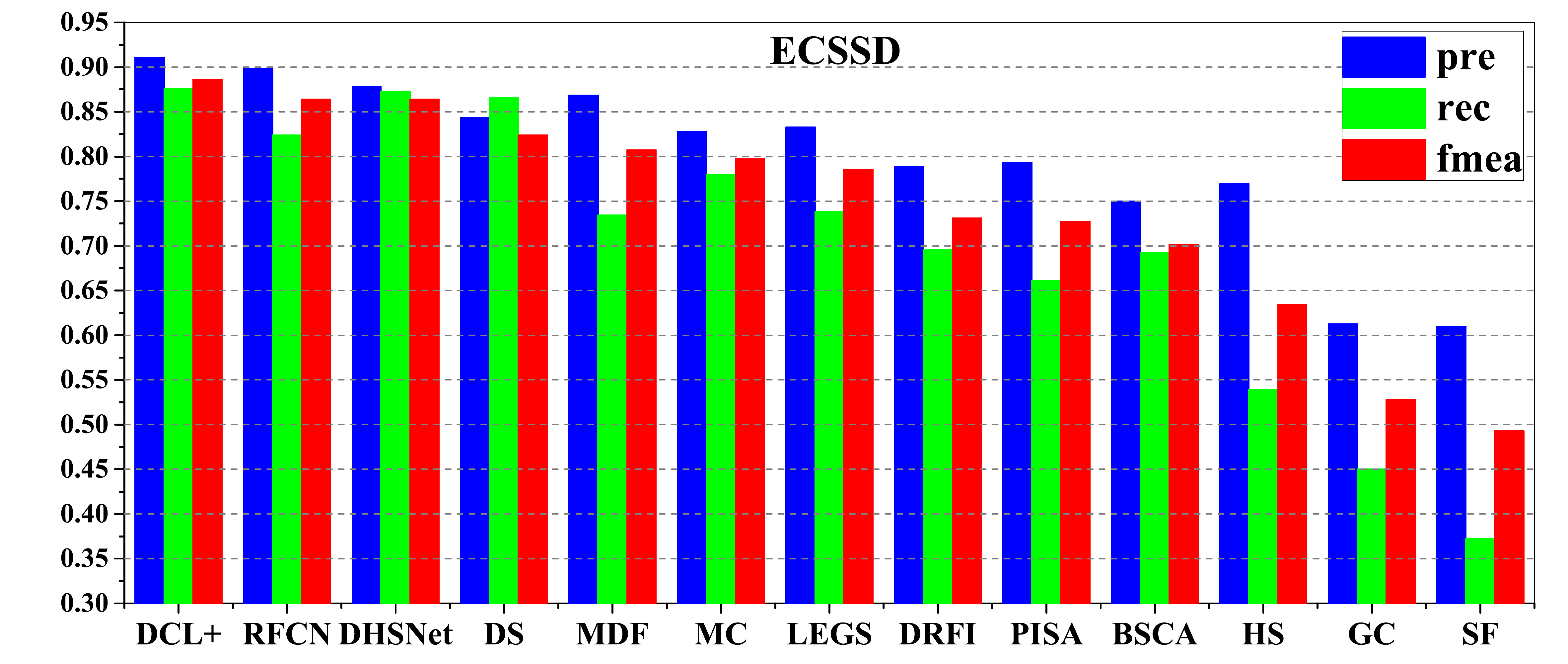}\hfill
    \includegraphics[width = 0.237\textwidth]{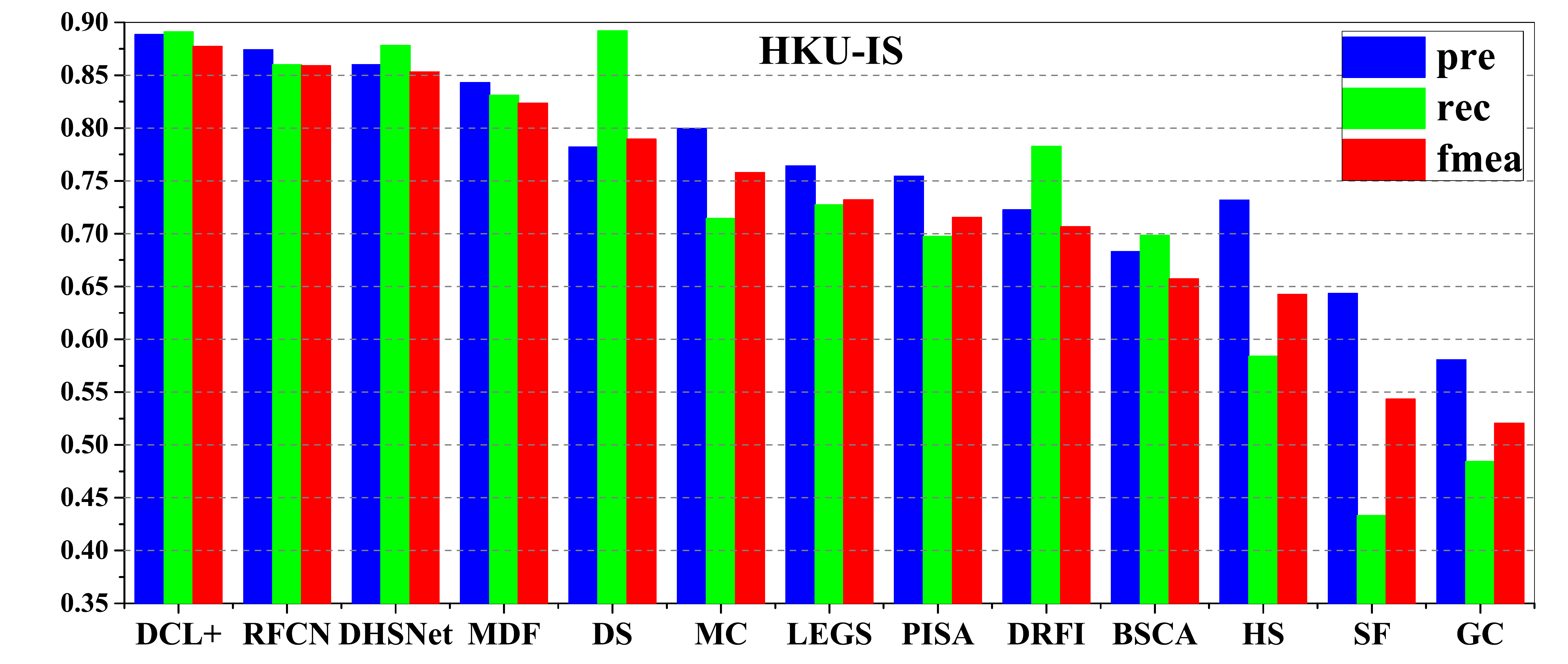}\hfill
    \includegraphics[width = 0.237\textwidth]{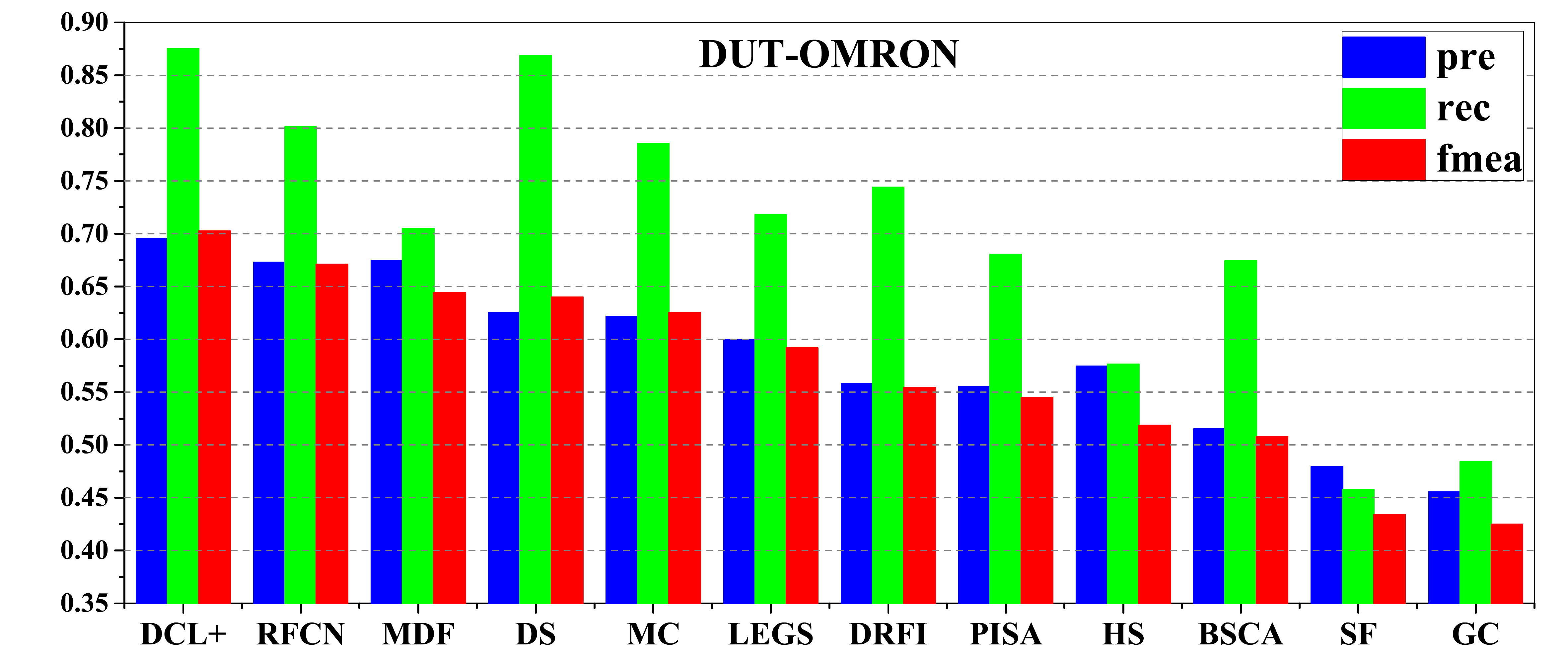}
  }
  \caption{Precision, recall and F-measure achieved using an adaptive threshold for every image. Our proposed method consistently performs best among 13 different methods on 4 datasets.
  }\label{fig:comps_prf}
\end{figure*}

\section{Experimental Results}\label{sec:experiment}
\subsection{Experimental Setup}
\subsubsection{Datasets}
We evaluate our proposed method on 6 widely used saliency detection benchmarks, including MSRA-B~\cite{liu2011learning}, HKU-IS~\cite{LiYu15}, PASCAL-S~\cite{li2014secrets}, DUT-OMRON~\cite{yang2013saliency}, ECSSD~\cite{yan2013hierarchical} and SOD~\cite{martin2001database}. MSRA-B includes 5,000 images, most of which holds a single salient object. HKU-IS is proposed in our previous work~\cite{LiYu15}, which has 4,447 images and most of the images include multiple separate salient objects. PASCAL-S is based on the validation set of PASCAL VOC2010 segmentation challenge~\cite{everingham2010pascal} and contains 850 natural images. DUT-OMRON has 5,168 challenging images, which have relatively complex and diversified contents. SOD has 300 images and was originally designed for image segmentation. It is very challenging as most of the images contain multiple objects and have low contrast or cluttered background. 
We train the proposed contrast-oriented deep neural networks based on the combination of both the training sets of the MSRA-B~(2500 images) and the HKU-IS~(2500 images). The two validation sets are also combined as our final validation, which contains a total of 1,000 images. 
We test the model trained on this combined training set over all other datasets to verity the model's adaptability.

\subsubsection{Evaluation Criteria}
We employ precision-recall~(PR) curves, F-measure and mean absolute error~(MAE)~to quantitatively evaluate the performance of our method as well as other salient object detection methods. Given a saliency map with continuous values normalized to the range of 0 and 255, we compute the binary masks by using every possible fixed integer threshold. A pair of precision/recall values can be computed by comparing each binary mask against the ground truth. The precision is defined as the ratio between detected groundtruth salient pixels and all predicted salient pixels in the binary mask while the recall being the ratio between detected groundtruth salient pixels and all groundtruth salient pixels. Once the precision/recall pairs of all binary maps have been computed, the PR curve can be plotted by averaging all pairs of precision and recall values over all saliency maps of a given dataset.
F-measure is defined as the harmonic mean of the average precision and the average recall, which can be calculated as
\begin{equation}
  F_{\beta} = \frac{(1+\beta^2)\cdot Precision \cdot Recall}{\beta^2\cdot Precision + Recall},
\end{equation}
where $\beta^2$ is set to 0.3 to place more emphasis on precision than recall, as suggested in~\cite{achanta2009frequency}. 
During evaluation, we report the maximum F-measure~(maxF) among all F-measure scores computed from precision/recall pairs on the PR curve. We also use twice the mean value of every saliency map as the threshold to generate the corresponding binary map and report the average precision, recall and F-measure of all binary maps. As a complement, we also calculate the mean absolute error~(MAE)~\cite{perazzi2012saliency} as follows to quantitatively measure the average absolute per-pixel difference between an estimated saliency map $S$ and the corresponding groundtruth saliency map $G$.
\begin{equation}
MAE = \frac{1}{W\times H}\sum_{x=1}^{W}\sum_{y=1}^{H}|S(x,y) - G(x,y)|.
\end{equation}

\subsubsection{Implementation}
Our proposed model has been implemented on top of the open source code of DeepLab~\cite{chen2014semantic}, which is based on the Caffe platform~\cite{jia2014caffe}. It was trained with a GTX Titan X GPU and Intel-i7 3.6GHz CPU.

During training, we resize all the images and their corresponding groundtruth saliency maps to $321\times 321$, and perform data augmentation by horizontal flipping. While training the MS-FCN stream, we set the learning rate for all newly added layers to $10^{-3}$ and the learning rate for the rest of the layers to $10^{-4}$. 
We employ a ``poly'' learning rate updating policy~\cite{liu2015parsenet} (the learning rate is scaled by $\left(1-\frac{iter}{max\_iter}\right)^{power}$ after each iteration, and $power = 0.9$). We set the weight decay to 0.0005 and the momentum parameter to 0.9 during training. For the segment-wise spatial pooling stream, we refer to~\cite{felzenszwalb2004efficient} and obtain 300 segments for each image from 3 levels of image segmentation achieved with different parameter settings. We set the grid size to $2\times 2$ while performing spatial pooling over each segment and the aggregated feature is of 6144 dimensions in the VGG-16 based MS-FCN and 12288 dimensions in the ResNet-101 based MS-FCN. 
This feature is further fed into a sub-network consisting of two fully connected layers, each of which contains 300 neurons. As in~\cite{krahenbuhl2012efficient}, we determine the parameters of the fully connected CRF by performing cross validation on the validation set. Finally, the actual value of $w_1$, $w_2$, $\sigma_\alpha$, $\sigma_\beta$, $\sigma_\gamma$ and $\sigma_\epsilon$ are respectively set to $3.0$, $5.0$, $3.0$, $50.0$, $3.0$ and $9.0$ during evaluation. 



We use DCL$^+$ and DCL to respectively represent our best saliency detectors with and without CRF-based refinement. While it takes approximately 25 hours to train our model, it only costs around 0.7 second for DCL to process an image of size $400\times300$ on a PC with NVIDIA Titan X GPU and Intel-i7 3.6GHz CPU. Note that this is far more efficient than region-wise deep saliency detectors which independently treat all image patches or superpixels during saliency estimation. 
However, CRF-based post-processing is more expensive, and requires additional 8 seconds since we need to compute generalized eigenvectors used in the CRF model. Experimental results reported in the following section show that DCL alone without CRF refinement already performs better than most of the existing state-of-the-art methods. A specific comparison of the computational cost of different methods is summarized in Table~\ref{tab:runtime_cmp}.

\begin{table*}[]
\centering
\resizebox{0.85\textwidth}{!}
{
\begin{tabular}{c|c|c|c|c|c|c|c|c|c|c|c|c|c}
\hline
        & SF    & GC   & HS   & DRFI  & PISA & BSCA & LEGS  & MC    & MDF   & DS    & RFCN  & DHSNet & DCL   \\ \hline
Time(s) & 0.115 & 0.25 & 0.43 & 47.08 & 0.65 & 2.03 & 2.00* & 2.38$^*$ & 8.00$^*$ & 0.25$^*$ & 4.60$^*$ & 0.24$^*$  & 0.68$^*$ \\ \hline
\end{tabular}
}
\caption{Comparison of running time. *: GPU time.}
\label{tab:runtime_cmp}
\end{table*}

\begin{table*}[]
\centering

\resizebox{0.95\textwidth}{!}
{
\begin{tabular}{|l|c|c|c|c|l|l|l|l|l|}
\hline
\multicolumn{4}{|c|}{MSFCN}                                                                                                                                                                     & \multicolumn{2}{c|}{Segment-Level}                                                                                                                  & \multicolumn{2}{c|}{CRF}                                          & \multicolumn{2}{c|}{Metric} \\ \hline
VGG16                            & \multicolumn{1}{l|}{ResNet-101} & \begin{tabular}[c]{@{}c@{}}Attentional\\ Module\end{tabular} & \begin{tabular}[c]{@{}c@{}}Multi-Scale\\ Input\end{tabular} & \begin{tabular}[c]{@{}c@{}}SLIC\\ Superpixel\end{tabular} & \multicolumn{1}{c|}{\begin{tabular}[c]{@{}c@{}}Multi-Scale\\ Segmentation\end{tabular}} & w/o contour                     & \multicolumn{1}{c|}{w/ contour} & maxF         & MAE          \\ \hline
\multicolumn{1}{|c|}{\textbf{${\surd}$}} & \multicolumn{1}{l|}{\textbf{}}  & \multicolumn{1}{l|}{\textbf{}}                               & \multicolumn{1}{l|}{\textbf{}}                              & \textbf{${\surd}$}                                                & \textbf{}                                                                               & \textbf{}                       & \textbf{}                       & 0.733        & 0.084        \\ \hline
\multicolumn{1}{|c|}{\textbf{${\surd}$}} & \multicolumn{1}{l|}{\textbf{}}  & \multicolumn{1}{l|}{\textbf{}}                               & \multicolumn{1}{l|}{\textbf{}}                              & \textbf{${\surd}$}                                                & \textbf{}                                                                               & \multicolumn{1}{c|}{\textbf{${\surd}$}} & \textbf{}                       & 0.757        & 0.080        \\ \hline
\multicolumn{1}{|c|}{\textbf{${\surd}$}} & \multicolumn{1}{l|}{\textbf{}}  & \textbf{${\surd}$}                                                   & \multicolumn{1}{l|}{\textbf{}}                              & \textbf{${\surd}$}                                                & \textbf{}                                                                               & \textbf{}                       & \textbf{}                       & 0.746        & 0.082        \\ \hline
\textbf{}                        & \textbf{${\surd}$}                      & \textbf{${\surd}$}                                                   & \multicolumn{1}{l|}{\textbf{}}                              & \textbf{${\surd}$}                                                & \textbf{}                                                                               & \textbf{}                       & \textbf{}                       & 0.773        & 0.076        \\ \hline
\textbf{}                        & \textbf{${\surd}$}                      & \textbf{${\surd}$}                                                   & \textbf{${\surd}$}                                                  & \textbf{${\surd}$}                                                & \textbf{}                                                                               & \textbf{}                       & \textbf{}                       & 0.792        & 0.071        \\ \hline
\textbf{}                        & \textbf{${\surd}$}                      & \textbf{${\surd}$}                                                   & \textbf{${\surd}$}                                                  & \multicolumn{1}{l|}{\textbf{}}                            & \multicolumn{1}{c|}{\textbf{${\surd}$}}                                                         & \textbf{}                       & \textbf{}                       & 0.799        & 0.070        \\ \hline
\textbf{}                        & \textbf{${\surd}$}                      & \textbf{${\surd}$}                                                   & \textbf{${\surd}$}                                                  & \textbf{}                                                 & \multicolumn{1}{c|}{\textbf{${\surd}$}}                                                         & \multicolumn{1}{c|}{\textbf{${\surd}$}} & \multicolumn{1}{c|}{\textbf{}}  & 0.804        & 0.068        \\ \hline
\textbf{}                        & \textbf{${\surd}$}                      & \textbf{${\surd}$}                                                   & \textbf{${\surd}$}                                                  & \textbf{}                                                 & \multicolumn{1}{c|}{\textbf{${\surd}$}}                                                         & \multicolumn{1}{c|}{\textbf{}}  & \multicolumn{1}{c|}{\textbf{${\surd}$}} & 0.811        & 0.064        \\ \hline
\end{tabular}
}
\caption{Performance evaluation of different model factors on DUT-OMRON Dataset.}
\label{tab:model_factor_analysis}
\end{table*}

\subsection{Comparison with the State of the Art}
We compare our models (DCL and DCL$^+$) with $9$ other state-of-the-art algorithms, including SF~\cite{perazzi2012saliency}, GC~\cite{cheng2015global}, HS~\cite{yan2013hierarchical}, DRFI~\cite{jiang2013salient}, PISA~\cite{pisa15PAMI}, BSCA~\cite{qin2015saliency}, LEGS~\cite{wang2015deep}, MC~\cite{zhao2015saliency}, MDF~\cite{LiYu15}, DS~\cite{li2015deepsaliency}, RFCN~\cite{wang2016saliency}, DHSNet~\cite{liu2016dhsnet}. The last three are fully convolutional neural network based methods, which were published after the publication of our earlier conference version~\cite{LiYu16}. 

For qualitative evaluation, Figure~\ref{fig:smaps} provides a visual comparison of saliency detection results, and the results from our proposed method achieve much improvement over those from other state-of-the-art algorithms. Specifically, our method is capable of highlighting salient regions missed by other methods in various challenging cases, e.g., salient regions touching the image boundary (the first and fifth rows), low contrast between salient objects and the background (the third and sixth rows) and images with multiple separate salient objects (the last three rows). 

Our method significantly outperforms all other methods, including those fully convolutional network based deep models published after our earlier conference version~\cite{LiYu16}, by a large margin on all public datasets in terms of the PR curve~(Fig.~\ref{fig:comps_pr}) as well as average precision, recall and F-measure (Fig.~\ref{fig:comps_prf}). 
Moreover, for the purpose of quantitative evaluation, we report a comparison of maximum F-measure and MAE in Table~\ref{tab:comp_quantity}. Our complete model~(DCL$^+$) clearly outperforms the previous best-performing method by 3.67\%, 1.98\%, 1.90\%, 10.64\%, 2.76\% and 3.38\% in terms of maximum F-measure on MSRA-B~(skipping RFCN and DHSNet on this dataset), ECSSD, HKU-IS, DUT-OMRON~(skipping DHSNet), PASCAL-S and SOD, respectively. And at the same time, it respectively lowers the MAE by 22.22\%, 1.69\%, 21.15\%, 23.81\%, 2.13\% and 5.51\%. 
It can also be observed that the proposed method~(DCL) without CRF-based post-processing already outperforms all evaluated methods on all considered datasets. We also compare run-time efficiency among the considered algorithms. As shown in Table~\ref{tab:runtime_cmp}, our DCL model needs around 0.68 second to generate a saliency map in the testing phase, which is comparable to other fully convolutional methods~(DS~\cite{li2015deepsaliency}, RFCN~\cite{wang2016saliency} and DHSNet~\cite{liu2016dhsnet}), and is much more efficient than other region-based CNN models~(LEGS~\cite{wang2015deep}, MC~\cite{zhao2015saliency}, MDF~\cite{LiYu15}).

\begin{figure}[ht]
    \centerline{
    \includegraphics[width = 0.225\textwidth]{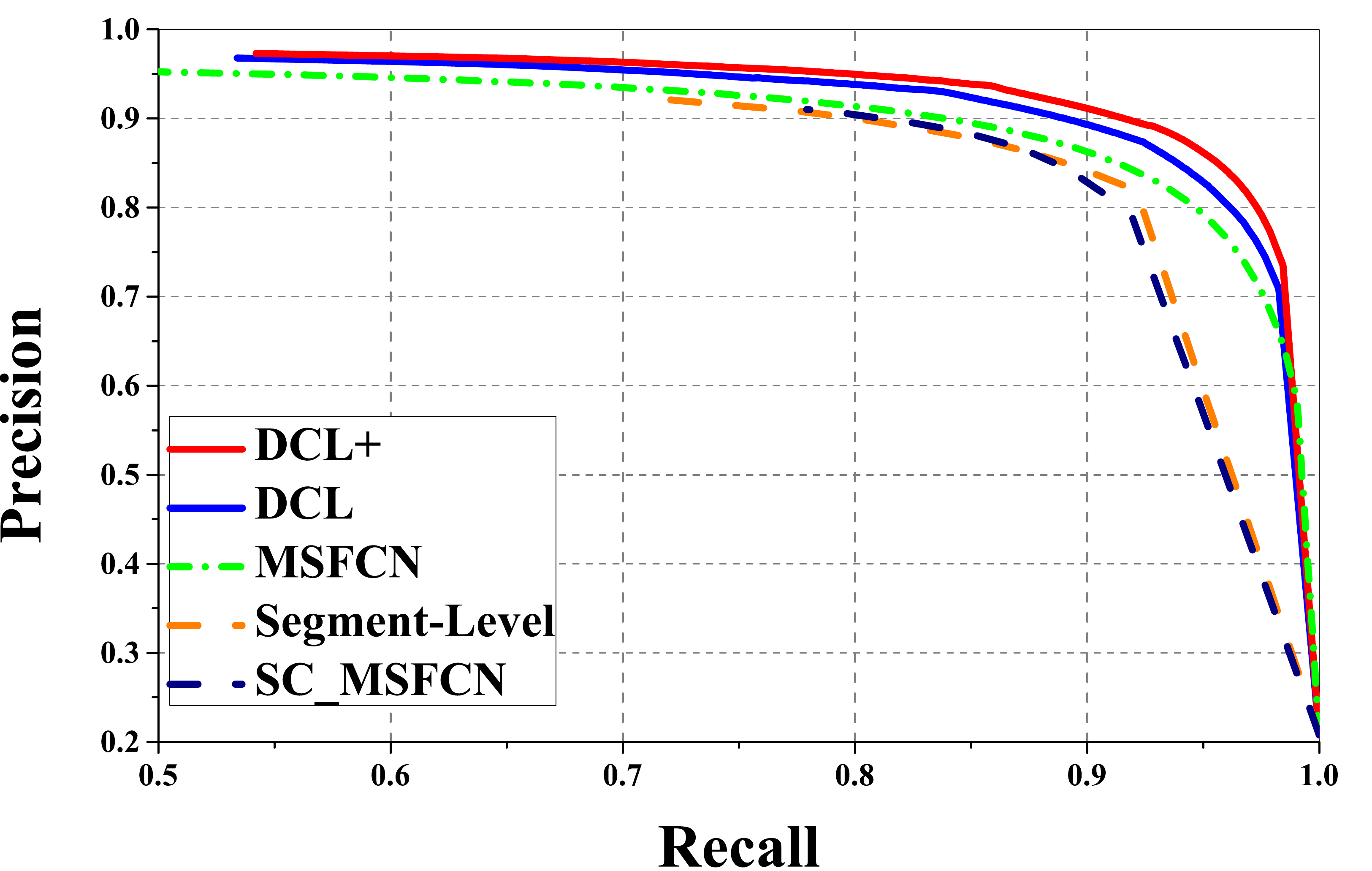}\hfill
    \includegraphics[width = 0.225\textwidth]{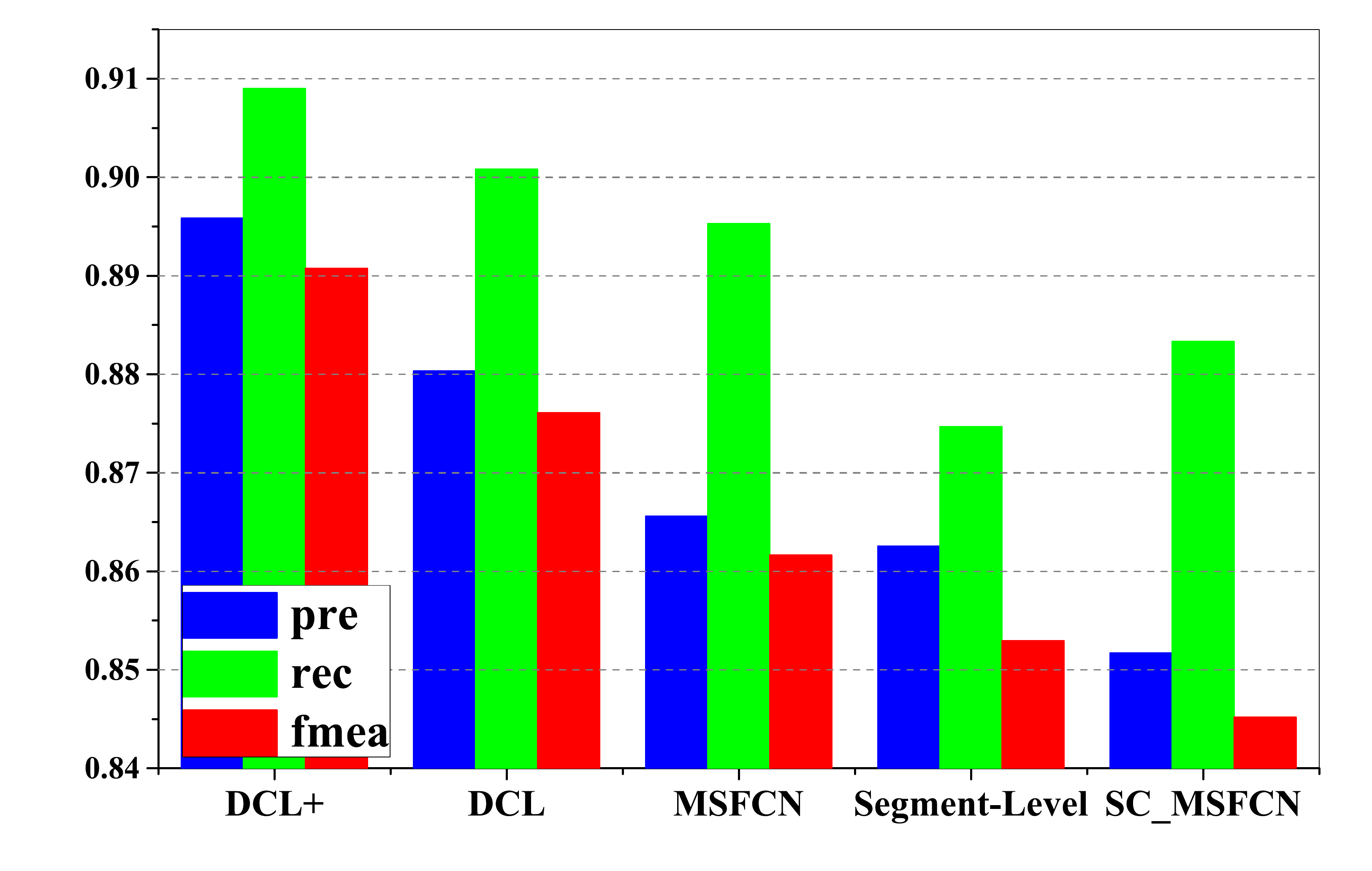}\hfill
  }\vspace{-0mm}
    \caption{Component-wise validation of the proposed model and the effectiveness of CRF based refinement.}
\label{fig:ablation_study}
\end{figure}

\begin{figure}[h]
\begin{center}
   \includegraphics[width=0.95\columnwidth]{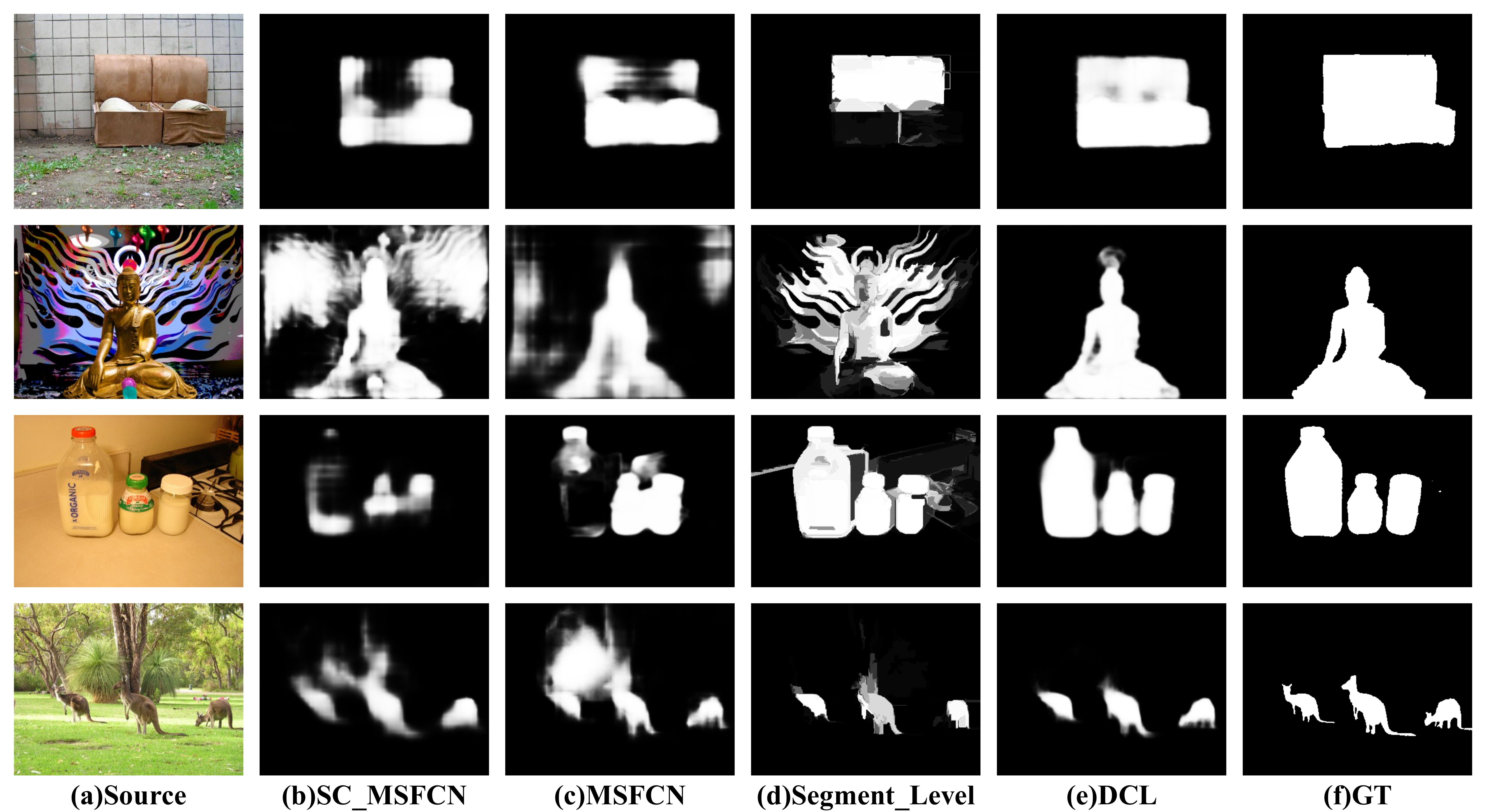}
\end{center}
   \caption{Sample visualizations demonstrating the component-wise efficacy of our deep contrast network.}
\label{fig:interp_cmp}
\end{figure}

\subsection{Ablation Studies}

\subsubsection{Component-wise Effectiveness of Deep Contrast Network}
To validate the necessity and effectiveness of the two components contained in our deep contrast network, we take the VGG-16 based version as a representative and compare the saliency maps~$S_1$ inferred from the first stream~(MS-FCN), the saliency maps $S_2$ from the second stream as well as the fused ones based on $S_1$ and $S_2$. As shown in Fig.~\ref{fig:ablation_study}, the fused saliency map consistently performs best under all evaluation metrics on the testing set of the MSRA-B dataset, and the fully convolutional stream contributes to the merged prediction far more than the segment-wise spatial pooling stream. The two streams of our deep contrast network are complementary and are capable of discovering global and local contrast collaboratively through multiscale feature aggregation in both streams. To validate the effectiveness of MS-FCN, we have also generated saliency maps from the last scale of MS-FCN for comparison. As illustrated in Fig.~\ref{fig:ablation_study}, a single scale of MS-FCN (SC\_MSFCN) may lead to significantly inferior performance when compared to the full version of MS-FCN in terms of the PR curve as well as average precision, recall and F-measure. Fig.~\ref{fig:interp_cmp} shows sample visualizations to demonstrate the complementary nature of the two streams inside the DCL network. As shown in the figure, although the fully convolutional stream and the segment-wise spatial pooling stream can produce promising saliency maps, they are far from perfect. MS-FCN tends to generate very smooth saliency maps but cannot well maintain the integrity of salient regions while the segment-wise stream predicts saliency maps in the unit of superpixels, it can hardly capture the global contrast and cannot well handle images with a complex background. However, the fused DCL model exploits the advantages of both and produces more accurate saliency predictions, which confirms the complementarity of these two sub-networks. In particular, there are examples~(e.g. the second image in Fig.~\ref{fig:interp_cmp}) where the two streams have different mistakenly predicted regions, but our proposed network still preferentially integrate respectively predicted salient pixels and produce more accurate results. This further demonstrates the robustness of our network and the strong complementarity of the two network streams.

\subsubsection{Effectiveness of Contour Guided CRF}\label{sec:effectiveness_spatial_coherence}
As described in Section~\ref{sec:spatial_coherence}, we incorporate a fully connected CRF with embedded contour features to further improve spatial coherence and contour positioning in the saliency maps generated from our deep contrast network. We compare the performance of the generated saliency maps with and without CRF as post-processing. As shown in Fig.~\ref{fig:ablation_study}, CRF significantly increases the accuracy of the saliency maps generated for the testing images of the MSRA-B dataset. We also show a visual comparison in Figure~\ref{fig:crf_effect} to illustrate the effectiveness of conventional CRF post-processing and CRF incorporating salient region contours. As shown in this figure, conventional CRF improves the spatial consistency of predicted results to a certain extent while incorporating salient region contours enhances the confidence of saliency predictions especially for pixels near detected salient region boundaries.

\subsection{Improvements after Conference Version}
After the conference version of this work, we have made the following five major modifications to our method: (1) adding an attention module to infer spatially varying weights for saliency map fusion, (2) employing the ResNet-101 network in the fully convolutional stream, (3) running the fully convolutional stream on multiple scaled versions of the original input image and fusing the results using max-pooling, (4) training and testing the segment-wise spatial pooling stream using segments from multi-level image segmentation, and (5) performing salient region contour detection and incorporating detected contours in the fully connected CRF during post-processing. In Table~\ref{tab:model_factor_analysis}, we evaluate how each of these factors affects the maximum F-measure and MAE on the DUT-OMRON dataset. As shown in the table, these five factors together contribute a 7.13\% improvement in the maximum F-measure and a 20.0\% decline in MAE in comparison to the best reported results in the earlier conference version of this paper.

\begin{figure}[b]
\begin{center}
   \includegraphics[width=0.95\columnwidth]{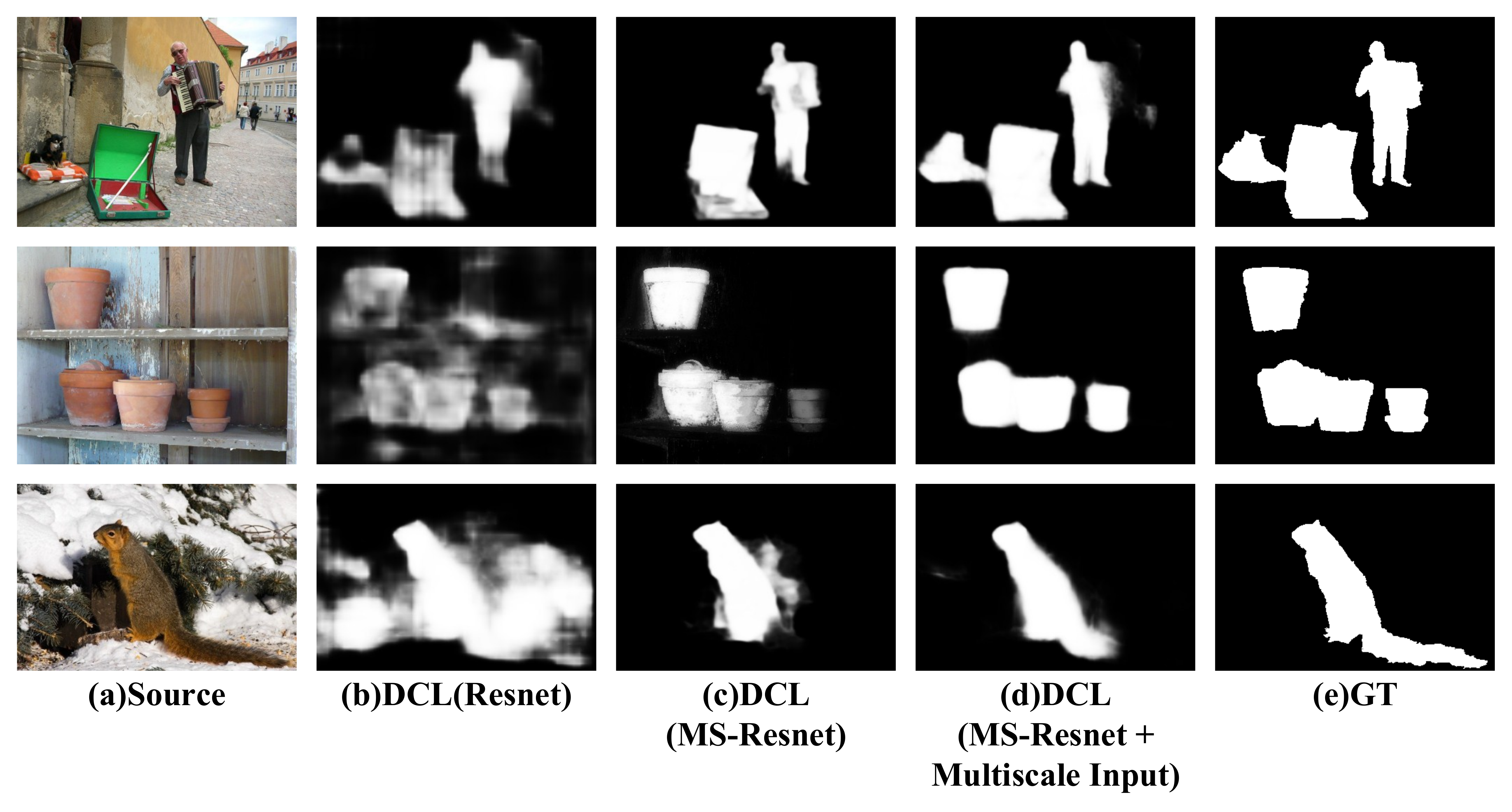}
\end{center}
   \caption{Effectiveness of ResNet-101 in our DCL model.}
\label{fig:resnet_effect}
\end{figure}

\subsubsection{Effectiveness of Attention Module}\label{effectiveness_atten}
As described in Section~\ref{sec:attentional_module}, instead of simply adding a $1\times 1$ convolutional layer on top of the saliency maps from the two network streams, we design an attention module to infer spatially varying weight maps. To validate its effectiveness, we conduct a performance comparison between a deep contrast network with a trained attention module and another deep contrast network with a simple $1\times 1$ convolutional layer. As shown in Table~\ref{tab:model_factor_analysis}, adopting the attention module for saliency map fusion improves the maximum F-measure on the DUT-OMRON dataset by 1.77\% while lowering the MAE by 2.38\%. Because of the effectiveness of this mechanism, we always integrate this module in our network in subsequent experiments.

\subsubsection{Effectiveness of ResNet-101 in MS-FCN}
As described in Section~\ref{sec:ms-fcn}, we have attempted to replace the VGG-16 network with a transformed ResNet-101 network in the fully convolutional stream of our deep network. To demonstrate its effectiveness, we have trained a new deep contrast network model for comparison. This new model is trained using the same setting as Section~\ref{effectiveness_atten} except that the transformed VGG-16 network is replaced with a pre-trained and transformed ResNet-101. As shown in Table~\ref{tab:model_factor_analysis}, adopting ResNet-101 instead of VGG-16 significantly improves the maximum F-measure on the DUT-OMRON dataset by 3.62\% while lowering the MAE by 7.32\%. We have also reached the same conclusion as the VGG based DCL network that ResNet-101 in the single scale setting generates over-smoothed saliency maps with prediction errors and performs much worse than the multi-scale version with side branches. As shown in the second and third columns of Fig~\ref{fig:resnet_effect}, our proposed DCL network with multi-scale ResNet-101 generates much more confident and cleaner results than DCL with the original single-scale ResNet-101.

\subsubsection{Effectiveness of Multiple Scaled Inputs}
Inspired by~\cite{chen2015attention}, we adopt a multi-scale input strategy when generating a saliency map from the fully convolutional stream. Specifically, we obtain three scaled versions of the original input image with the scaling factor respectively set to $1$, $0.75$, and $0.5$, and independently feed these scaled images to the fully convolutional stream. The three resulting saliency maps are fused by taking the maximum response across scales for each position (i.e. max pooling). As shown in Table~\ref{tab:model_factor_analysis}, multi-scale input brings an extra 2.46\% improvement in the maximum F-measure while lowering the MAE by 6.58\%. Sample visualizations are shown in the fourth column of Fig~\ref{fig:resnet_effect}, where fusing saliency predictions from multi-scale inputs gives rise to more accurate saliency maps especially when there exists multiple salient objects of different scales in the testing image.

\subsubsection{Effectiveness of Multi-Level Image Segmentation}
As described in Section~\ref{sec:superpixel}, the final saliency map from the revised segment-wise spatial pooling stream is the average of three saliency maps, each of which is computed using all superpixels from one of $3$ levels of image segmentation. As shown in Table~\ref{tab:model_factor_analysis}, multi-level image segmentation further improves the maximum F-measure by 0.88\% and lowers the MAE by 1.40\%.

\subsubsection{Effectiveness of Salient Region Contours}
As described in Section~\ref{sec:algo}, we revise the CRF-based post-processing step in this version by integrating an additional feature vector computed from detected salient region contours. Salient region contours are detected using a separately trained contour detection model, which has the same network structure as the MS-FCN stream. We compare saliency maps computed without CRF, with CRF but without contour saliency features, and with contour guided CRF, respectively. As shown in Table~\ref{tab:model_factor_analysis}, post-processing our saliency maps with a dense CRF always yields performance improvement. For the VGG16 based deep contrast network, running CRF as a post-processing step boosts the maximum F-measure by 3.27\% and lowers the MAE by 4.76\%. For the ResNet-101 based deep contrast network, which already achieves a much better performance itself, adding a dense CRF still brings a 0.63\% improvement in the maximum F-measure and a 2.86\% decrease in MAE. It is worth noting that contour guided CRF results in more accurate saliency maps with a 1.50\% improvement in the maximum F-measure and a 8.57\% decrease in MAE.


\section{Conclusions}\label{sec:conclusion}

In this work, we have proposed end-to-end contrast-oriented deep neural networks for salient object detection. Our deep networks contain two complementary sub-networks and are capable of extracting a wide variety of visual contrast information. The first sub-network is based on a multiscale fully convolutional network, and is intended to infer pixel-wise saliency by looking into contexts~(receptive field) of multiple scales around each pixel. The second sub-network is designed to capture the contrast information among adjacent regions, which can not only maintain the consistency of saliency prediction within homogeneous regions but also better detect discontinuities along salient region boundaries. An attentional module with learnable weights is introduced to adaptively fuse the two saliency maps from the two sub-networks. Finally, to produce more accurate saliency predictions, we incorporate a CRF with a contour feature embedding to further enhance the spatial coherence and contour localization of the produced saliency map. Experimental results show that the proposed model achieves state-of-the-art performance on six public benchmark datasets under various evaluation metrics.

\ifCLASSOPTIONcaptionsoff
  \newpage
\fi



%



 {\small

 \bibliography{dcl_saliency}{}

\begin{thebibliography}{10}
\providecommand{\url}[1]{#1}
\csname url@samestyle\endcsname
\providecommand{\newblock}{\relax}
\providecommand{\bibinfo}[2]{#2}
\providecommand{\BIBentrySTDinterwordspacing}{\spaceskip=0pt\relax}
\providecommand{\BIBentryALTinterwordstretchfactor}{4}
\providecommand{\BIBentryALTinterwordspacing}{\spaceskip=\fontdimen2\font plus
\BIBentryALTinterwordstretchfactor\fontdimen3\font minus
  \fontdimen4\font\relax}
\providecommand{\BIBforeignlanguage}[2]{{%
\expandafter\ifx\csname l@#1\endcsname\relax
\typeout{** WARNING: IEEEtran.bst: No hyphenation pattern has been}%
\typeout{** loaded for the language `#1'. Using the pattern for}%
\typeout{** the default language instead.}%
\else
\language=\csname l@#1\endcsname
\fi
#2}}
\providecommand{\BIBdecl}{\relax}
\BIBdecl

\bibitem{LiYu16}
G.~Li and Y.~Yu, ``Deep contrast learning for salient object detection,'' in
  \emph{Proc. IEEE Conf. CVPR}, June 2016.

\bibitem{wei2016stc}
Y.~Wei, X.~Liang, Y.~Chen, X.~Shen, M.-M. Cheng, J.~Feng, Y.~Zhao, and S.~Yan,
  ``Stc: A simple to complex framework for weakly-supervised semantic
  segmentation,'' \emph{IEEE Trans. Pattern Anal. Mach. Intell.}, 2016.

\bibitem{navalpakkam2006integrated}
V.~Navalpakkam and L.~Itti, ``An integrated model of top-down and bottom-up
  attention for optimizing detection speed,'' in \emph{Proc. IEEE Conf. CVPR},
  vol.~2, 2006, pp. 2049--2056.

\bibitem{wang2017multi}
Z.~Wang, T.~Chen, G.~Li, R.~Xu, and L.~Lin, ``Multi-label image recognition by
  recurrently discovering attentional regions,'' in \emph{Proc. IEEE Conf.
  ICCV}, 2017.

\bibitem{avidan2007seam}
S.~Avidan and A.~Shamir, ``Seam carving for content-aware image resizing,'' in
  \emph{ACM Transactions on graphics (TOG)}, vol.~26, no.~3.\hskip 1em plus
  0.5em minus 0.4em\relax ACM, 2007, p.~10.

\bibitem{wu2014weighted}
H.~Wu, G.~Li, and X.~Luo, ``Weighted attentional blocks for probabilistic
  object tracking,'' \emph{The Visual Computer}, vol.~30, no.~2, pp. 229--243,
  2014.

\bibitem{bi2014person}
S.~Bi, G.~Li, and Y.~Yu, ``Person re-identification using multiple experts with
  random subspaces,'' \emph{Journal of Image and Graphics}, vol.~2, no.~2,
  2014.

\bibitem{einhauser2003does}
W.~Einh{\"a}user and P.~Ko{\`E}nig, ``Does luminance-contrast contribute to a
  saliency map for overt visual attention?'' \emph{European Journal of
  Neuroscience}, vol.~17, no.~5, pp. 1089--1097, 2003.

\bibitem{parkhurst2002modeling}
D.~Parkhurst, K.~Law, and E.~Niebur, ``Modeling the role of salience in the
  allocation of overt visual attention,'' \emph{Vision research}, vol.~42,
  no.~1, pp. 107--123, 2002.

\bibitem{cheng2015global}
M.~Cheng, N.~J. Mitra, X.~Huang, P.~H. Torr, and S.~Hu, ``Global contrast based
  salient region detection,'' \emph{IEEE Trans. Pattern Anal. Mach. Intell.},
  vol.~37, no.~3, pp. 569--582, 2015.

\bibitem{yang2013saliency}
C.~Yang, L.~Zhang, H.~Lu, X.~Ruan, and M.-H. Yang, ``Saliency detection via
  graph-based manifold ranking,'' in \emph{Proc. IEEE Conf. CVPR}, 2013, pp.
  3166--3173.

\bibitem{wang2013visual}
Q.~Wang, Y.~Yuan, and P.~Yan, ``Visual saliency by selective contrast,''
  \emph{IEEE Transactions on Circuits and Systems for Video Technology},
  vol.~23, no.~7, pp. 1150--1155, 2013.

\bibitem{lu2014learning}
S.~Lu, V.~Mahadevan, and N.~Vasconcelos, ``Learning optimal seeds for
  diffusion-based salient object detection,'' in \emph{Proc. IEEE Conf. CVPR},
  2014, pp. 2790--2797.

\bibitem{jiang2013salient}
P.~Jiang, H.~Ling, J.~Yu, and J.~Peng, ``Salient region detection by ufo:
  Uniqueness, focusness and objectness,'' in \emph{Proc. IEEE Conf. ICCV},
  2013, pp. 1976--1983.

\bibitem{liu2011learning}
T.~Liu, Z.~Yuan, J.~Sun, J.~Wang, N.~Zheng, X.~Tang, and H.-Y. Shum, ``Learning
  to detect a salient object,'' \emph{IEEE Trans. Pattern Anal. Mach. Intell.},
  vol.~33, no.~2, pp. 353--367, 2011.

\bibitem{mai2013saliency}
L.~Mai, Y.~Niu, and F.~Liu, ``Saliency aggregation: a data-driven approach,''
  in \emph{Proc. IEEE Conf. CVPR}, 2013, pp. 1131--1138.

\bibitem{LiYu15}
G.~Li and Y.~Yu, ``Visual saliency based on multiscale deep features,'' in
  \emph{Proc. IEEE Conf. CVPR}, June 2015.

\bibitem{zhao2015saliency}
R.~Zhao, W.~Ouyang, H.~Li, and X.~Wang, ``Saliency detection by multi-context
  deep learning,'' in \emph{Proc. IEEE Conf. CVPR}, 2015, pp. 1265--1274.

\bibitem{wang2015deep}
L.~Wang, H.~Lu, X.~Ruan, and M.-H. Yang, ``Deep networks for saliency detection
  via local estimation and global search,'' in \emph{Proc. IEEE Conf. CVPR},
  2015, pp. 3183--3192.

\bibitem{long2014fully}
J.~Long, E.~Shelhamer, and T.~Darrell, ``Fully convolutional networks for
  semantic segmentation,'' \emph{Proc. IEEE Conf. CVPR}, 2015.

\bibitem{chen2014semantic}
L.-C. Chen, G.~Papandreou, I.~Kokkinos, K.~Murphy, and A.~L. Yuille, ``Semantic
  image segmentation with deep convolutional nets and fully connected crfs,''
  \emph{arXiv preprint arXiv:1412.7062}, 2014.

\bibitem{xie2015holistically}
S.~Xie and Z.~Tu, ``Holistically-nested edge detection,'' \emph{Proc. IEEE
  Conf. ICCV}, 2015.

\bibitem{gao2007bottom}
D.~Gao and N.~Vasconcelos, ``Bottom-up saliency is a discriminant process,'' in
  \emph{Proc. IEEE Conf. ICCV}, 2007, pp. 1--6.

\bibitem{achanta2009frequency}
R.~Achanta, S.~Hemami, F.~Estrada, and S.~Susstrunk, ``Frequency-tuned salient
  region detection,'' in \emph{Proc. IEEE Conf. CVPR}, 2009, pp. 1597--1604.

\bibitem{klein2011center}
D.~Klein, S.~Frintrop \emph{et~al.}, ``Center-surround divergence of feature
  statistics for salient object detection,'' in \emph{Proc. IEEE Conf.
  ICCV}.\hskip 1em plus 0.5em minus 0.4em\relax IEEE, 2011, pp. 2214--2219.

\bibitem{perazzi2012saliency}
F.~Perazzi, P.~Kr{\"a}henb{\"u}hl, Y.~Pritch, and A.~Hornung, ``Saliency
  filters: Contrast based filtering for salient region detection,'' in
  \emph{Proc. IEEE Conf. CVPR}, 2012, pp. 733--740.

\bibitem{zhu2014saliency}
W.~Zhu, S.~Liang, Y.~Wei, and J.~Sun, ``Saliency optimization from robust
  background detection,'' in \emph{Proc. IEEE Conf. CVPR}, 2014, pp.
  2814--2821.

\bibitem{wang2013saliency}
Q.~Wang, Y.~Yuan, P.~Yan, and X.~Li, ``Saliency detection by multiple-instance
  learning,'' \emph{IEEE transactions on cybernetics}, vol.~43, no.~2, pp.
  660--672, 2013.

\bibitem{judd2009learning}
T.~Judd, K.~Ehinger, F.~Durand, and A.~Torralba, ``Learning to predict where
  humans look,'' in \emph{Proc. IEEE Conf. ICCV}, 2009.

\bibitem{chang2011fusing}
K.-Y. Chang, T.-L. Liu, H.-T. Chen, and S.-H. Lai, ``Fusing generic objectness
  and visual saliency for salient object detection,'' in \emph{Proc. IEEE Conf.
  ICCV}.\hskip 1em plus 0.5em minus 0.4em\relax IEEE, 2011, pp. 914--921.

\bibitem{goferman2012context}
S.~Goferman, L.~Zelnik-Manor, and A.~Tal, ``Context-aware saliency detection,''
  \emph{TPAMI}, vol.~34, no.~10, pp. 1915--1926, 2012.

\bibitem{shen2012unified}
X.~Shen and Y.~Wu, ``A unified approach to salient object detection via low
  rank matrix recovery,'' in \emph{Proc. IEEE Conf. CVPR}, 2012.

\bibitem{liu2014adaptive}
R.~Liu, J.~Cao, Z.~Lin, and S.~Shan, ``Adaptive partial differential equation
  learning for visual saliency detection,'' in \emph{Proc. IEEE Conf. CVPR},
  2014.

\bibitem{jia2013category}
Y.~Jia and M.~Han, ``Category-independent object-level saliency detection,'' in
  \emph{Proc. IEEE Conf. ICCV}.\hskip 1em plus 0.5em minus 0.4em\relax IEEE,
  2013, pp. 1761--1768.

\bibitem{li2014secrets}
Y.~Li, X.~Hou, C.~Koch, J.~M. Rehg, and A.~L. Yuille, ``The secrets of salient
  object segmentation,'' in \emph{Proc. IEEE Conf. CVPR}, 2014, pp. 280--287.

\bibitem{hou2007saliency}
X.~Hou and L.~Zhang, ``Saliency detection: A spectral residual approach,'' in
  \emph{Proc. IEEE Conf. CVPR}, 2007.

\bibitem{lei2016universal}
J.~Lei, B.~Wang, Y.~Fang, W.~Lin, P.~Le~Callet, N.~Ling, and C.~Hou, ``A
  universal framework for salient object detection,'' \emph{IEEE Transactions
  on Multimedia}, vol.~18, no.~9, pp. 1783--1795, 2016.

\bibitem{krizhevsky2012imagenet}
A.~Krizhevsky, I.~Sutskever, and G.~E. Hinton, ``Imagenet classification with
  deep convolutional neural networks,'' in \emph{Advances in neural information
  processing systems}, 2012, pp. 1097--1105.

\bibitem{girshick2014rich}
R.~Girshick, J.~Donahue, T.~Darrell, and J.~Malik, ``Rich feature hierarchies
  for accurate object detection and semantic segmentation,'' in \emph{Proc.
  IEEE Conf. CVPR}, 2014, pp. 580--587.

\bibitem{li2016visual}
G.~Li and Y.~Yu, ``Visual saliency detection based on multiscale deep cnn
  features,'' \emph{IEEE Transactions on Image Processing}, vol.~25, no.~11,
  pp. 5012--5024, 2016.

\bibitem{li2015deepsaliency}
X.~Li, L.~Zhao, L.~Wei, M.~Yang, F.~Wu, Y.~Zhuang, H.~Ling, and J.~Wang,
  ``Deepsaliency: Multi-task deep neural network model for salient object
  detection,'' \emph{arXiv preprint arXiv:1510.05484}, 2015.

\bibitem{li2017instance}
G.~Li, Y.~Xie, L.~Lin, and Y.~Yu, ``Instance-level salient object
  segmentation,'' \emph{Proc. IEEE Conf. CVPR}, 2017.

\bibitem{LiAAAI18}
G.~Li, Y.~Xie, and L.~Lin, ``Weakly supervised salient object detection using
  image labels,'' in \emph{Proc. Conf. AAAI}, Feb 2018.

\bibitem{han2016two}
J.~Han, D.~Zhang, S.~Wen, L.~Guo, T.~Liu, and X.~Li, ``Two-stage learning to
  predict human eye fixations via sdaes,'' \emph{IEEE transactions on
  cybernetics}, vol.~46, no.~2, pp. 487--498, 2016.

\bibitem{li2015weighted}
N.~Li, B.~Sun, and J.~Yu, ``A weighted sparse coding framework for saliency
  detection,'' in \emph{Proc. IEEE Conf. CVPR}, 2015, pp. 5216--5223.

\bibitem{liu2016dhsnet}
N.~Liu and J.~Han, ``Dhsnet: Deep hierarchical saliency network for salient
  object detection,'' in \emph{Proc. IEEE Conf. CVPR}, 2016, pp. 678--686.

\bibitem{kuen2016recurrent}
J.~Kuen, Z.~Wang, and G.~Wang, ``Recurrent attentional networks for saliency
  detection,'' \emph{arXiv preprint arXiv:1604.03227}, 2016.

\bibitem{wang2016saliency}
L.~Wang, L.~Wang, H.~Lu, P.~Zhang, and X.~Ruan, ``Saliency detection with
  recurrent fully convolutional networks,'' in \emph{Proc. Conf. ECCV}.\hskip
  1em plus 0.5em minus 0.4em\relax Springer, 2016, pp. 825--841.

\bibitem{he2015deep}
K.~He, X.~Zhang, S.~Ren, and J.~Sun, ``Deep residual learning for image
  recognition,'' \emph{arXiv preprint arXiv:1512.03385}, 2015.

\bibitem{simonyan2014very}
K.~Simonyan and A.~Zisserman, ``Very deep convolutional networks for
  large-scale image recognition,'' \emph{arXiv preprint arXiv:1409.1556}, 2014.

\bibitem{li2014highly}
H.~Li, R.~Zhao, and X.~Wang, ``Highly efficient forward and backward
  propagation of convolutional neural networks for pixelwise classification,''
  \emph{arXiv preprint arXiv:1412.4526}, 2014.

\bibitem{mallat1999wavelet}
S.~Mallat, \emph{A wavelet tour of signal processing}.\hskip 1em plus 0.5em
  minus 0.4em\relax Academic press, 1999.

\bibitem{girshickICCV15fastrcnn}
R.~Girshick, ``Fast r-cnn,'' in \emph{International Conference on Computer
  Vision ({ICCV})}, 2015.

\bibitem{he2014spatial}
K.~He, X.~Zhang, S.~Ren, and J.~Sun, ``Spatial pyramid pooling in deep
  convolutional networks for visual recognition,'' in \emph{Proc. Conf.
  ECCV}.\hskip 1em plus 0.5em minus 0.4em\relax Springer, 2014, pp. 346--361.

\bibitem{bahdanau2014neural}
D.~Bahdanau, K.~Cho, and Y.~Bengio, ``Neural machine translation by jointly
  learning to align and translate,'' \emph{Proc. Conf. ICLR}, 2015.

\bibitem{chen2015attention}
L.-C. Chen, Y.~Yang, J.~Wang, W.~Xu, and A.~L. Yuille, ``Attention to scale:
  Scale-aware semantic image segmentation,'' \emph{arXiv preprint
  arXiv:1511.03339}, 2015.

\bibitem{deng2009imagenet}
J.~Deng, W.~Dong, R.~Socher, L.-J. Li, K.~Li, and L.~Fei-Fei, ``Imagenet: A
  large-scale hierarchical image database,'' in \emph{Proc. IEEE Conf. CVPR},
  2009, pp. 248--255.

\bibitem{criminisi2010geodesic}
A.~Criminisi, T.~Sharp, C.~Rother, and P.~P{\'e}rez, ``Geodesic image and video
  editing.'' \emph{ACM Transactions on graphics (TOG)}.

\bibitem{felzenszwalb2004efficient}
P.~F. Felzenszwalb and D.~P. Huttenlocher, ``Efficient graph-based image
  segmentation,'' \emph{IJCV}, vol.~59, no.~2, pp. 167--181, 2004.

\bibitem{shi2000normalized}
J.~Shi and J.~Malik, ``Normalized cuts and image segmentation,'' \emph{IEEE
  Trans. Pattern Anal. Mach. Intell.}, vol.~22, no.~8, pp. 888--905, 2000.

\bibitem{krahenbuhl2012efficient}
P.~Kr{\"a}henb{\"u}hl and V.~Koltun, ``Efficient inference in fully connected
  crfs with gaussian edge potentials,'' \emph{arXiv preprint arXiv:1210.5644},
  2012.

\bibitem{shotton2009textonboost}
J.~Shotton, J.~Winn, C.~Rother, and A.~Criminisi, ``Textonboost for image
  understanding: Multi-class object recognition and segmentation by jointly
  modeling texture, layout, and context,'' \emph{International Journal of
  Computer Vision}, vol.~81, no.~1, pp. 2--23, 2009.

\bibitem{yan2013hierarchical}
Q.~Yan, L.~Xu, J.~Shi, and J.~Jia, ``Hierarchical saliency detection,'' in
  \emph{Proc. IEEE Conf. CVPR}, 2013.

\bibitem{martin2001database}
D.~Martin, C.~Fowlkes, D.~Tal, and J.~Malik, ``A database of human segmented
  natural images and its application to evaluating segmentation algorithms and
  measuring ecological statistics,'' in \emph{Proc. IEEE Conf. ICCV}, vol.~2,
  2001, pp. 416--423.

\bibitem{everingham2010pascal}
M.~Everingham, L.~Van~Gool, C.~K. Williams, J.~Winn, and A.~Zisserman, ``The
  pascal visual object classes (voc) challenge,'' \emph{International journal
  of computer vision}, vol.~88, no.~2, pp. 303--338, 2010.

\bibitem{jia2014caffe}
Y.~Jia, E.~Shelhamer, J.~Donahue, S.~Karayev, J.~Long, R.~Girshick,
  S.~Guadarrama, and T.~Darrell, ``Caffe: Convolutional architecture for fast
  feature embedding,'' in \emph{Proceedings of the ACM International Conference
  on Multimedia}.\hskip 1em plus 0.5em minus 0.4em\relax ACM, 2014, pp.
  675--678.

\bibitem{liu2015parsenet}
W.~Liu, A.~Rabinovich, and A.~C. Berg, ``Parsenet: Looking wider to see
  better,'' \emph{arXiv preprint arXiv:1506.04579}, 2015.

\bibitem{pisa15PAMI}
K.~Wang, L.~Lin, J.~Lu, C.~Li, and K.~Shi, ``Pisa: Pixelwise image saliency by
  aggregating complementary appearance contrast measures with edge-preserving
  coherence,'' \emph{IEEE Transactions on Image Processing}, vol.~24, no.~10,
  pp. 3019--3033, Oct 2015.

\bibitem{qin2015saliency}
Y.~Qin, H.~Lu, Y.~Xu, and H.~Wang, ``Saliency detection via cellular
  automata,'' in \emph{Proc. IEEE Conf. CVPR}, 2015, pp. 110--119.

\end{thebibliography}
 \bibliographystyle{IEEEtran}
 }

%

\begin{IEEEbiography}[{\includegraphics[width=1in,height=1.25in,clip,keepaspectratio]{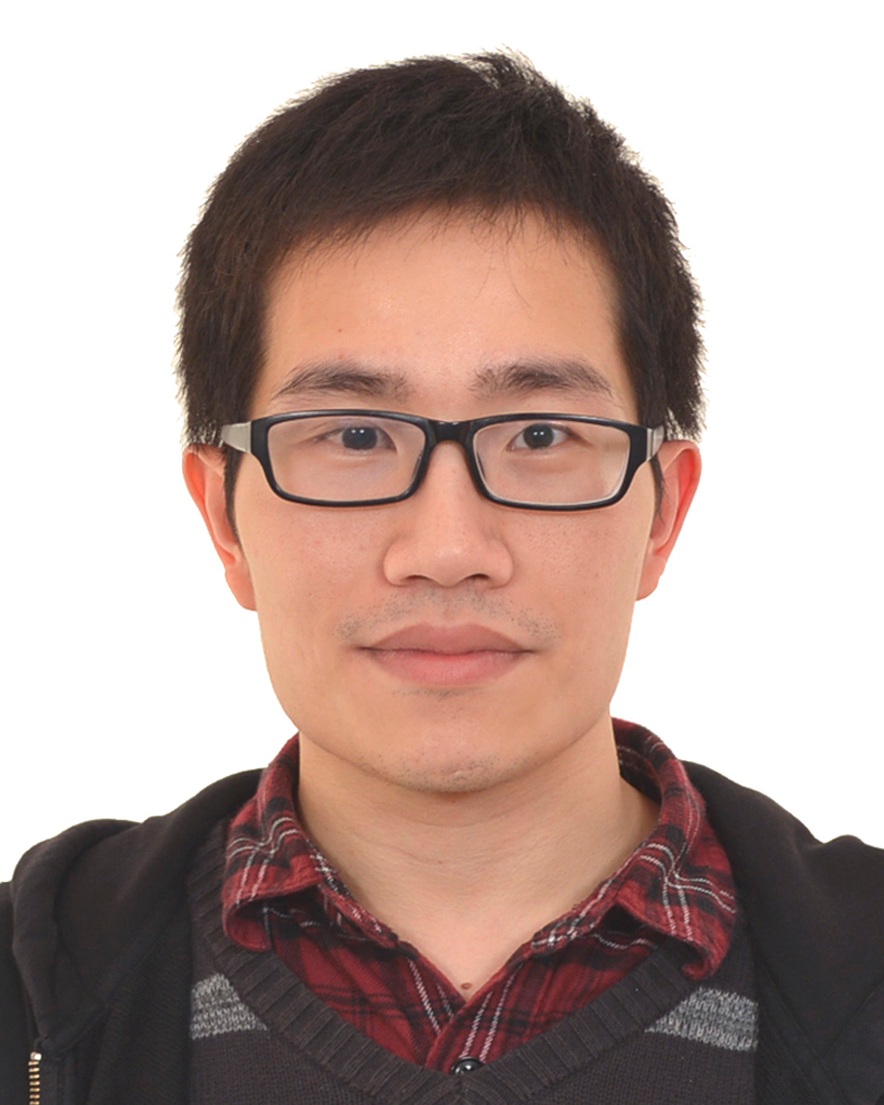}}]{Guanbin Li} is currently a research associate professor in School of Data and Computer Science, Sun Yat-sen University. He received his PhD degree from the University of Hong Kong in 2016. He was a recipient of Hong Kong Postgraduate Fellowship. His current research interests include computer vision, image processing, and deep learning. He has authorized and co-authorized on more than 20 papers in top-tier academic journals and conferences. He serves as an area chair for the conference of VISAPP. He has been serving as a reviewer for  numerous academic journals and conferences such as TPAMI, TIP, TMM, TC, CVPR2018 and IJCAI2018.
\end{IEEEbiography}

\begin{IEEEbiography}[{\includegraphics[width=1in,height=1.25in,clip,keepaspectratio]{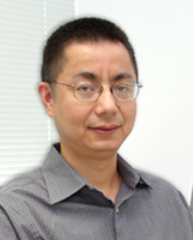}}]{Yizhou Yu} received the PhD degree from University of California at Berkeley in 2000. He is currently a professor at The University of Hong Kong, and was a faculty member at University of Illinois, Urbana-Champaign between 2000 and 2012. He is a recipient of 2002 US National Science Foundation CAREER Award, and 2007 NNSF China Overseas Distinguished Young Investigator Award. He has served on the editorial board of IET Computer Vision, IEEE Transactions on Visualization and Computer Graphics, The Visual Computer, and International Journal of Software and Informatics. He has also served on the program committee of many leading international conferences, including SIGGRAPH, SIGGRAPH Asia, and International Conference on Computer Vision. His current research interests include deep learning methods for computer vision, computational visual media, geometric computing, video analytics and biomedical data analysis.
\end{IEEEbiography}








\end{document}